\pdfoutput=1
\documentclass[11pt]{article}

\usepackage[]{ACL2023}
\usepackage{times}
\usepackage{latexsym}
\usepackage[T1]{fontenc}
\usepackage[utf8]{inputenc}
\usepackage{microtype}
\usepackage{inconsolata}
\usepackage[mathscr]{euscript}
\usepackage{color}
\usepackage{graphicx}
\usepackage{makecell}
\usepackage{color,soul}
\usepackage{amsmath}
\usepackage{enumitem}
\usepackage{csquotes}
\usepackage{booktabs}
\usepackage{xcolor}
\usepackage{hyperref}
\newcolumntype{?}{!{\vrule width 1pt}}

\newcommand{\methodname}[0]{$T^5Score$}
\newcommand{\ldalike}{WD}
\newcommand{\llmbased}{FT}

\DeclareSymbolFont{rsfs}{U}{rsfs}{m}{n}
\DeclareSymbolFontAlphabet{\mathscrsfs}{rsfs}

\title{\methodname: A Methodology for Automatically Assessing the Quality of LLM Generated Multi-Document Topic Sets}

\author{Itamar Trainin \quad Omri Abend \\
        The Hebrew University of Jerusalem \\
        \texttt{\{first\}.\{last\}@mail.huji.ac.il}}

\begin{document}
\maketitle

\begin{abstract} 

Using LLMs for Multi-Document Topic Extraction has recently gained popularity due to their apparent high-quality outputs, expressiveness, and ease of use.
However, most existing evaluation practices are not designed for LLM-generated topics and result in low inter-annotator agreement scores, hindering the reliable use of LLMs for the task.
To address this, we introduce \methodname, an evaluation methodology that decomposes the quality of a topic set into quantifiable aspects, measurable through easy-to-perform annotation tasks. 
This framing enables a convenient, manual or automatic, evaluation procedure resulting in a strong inter-annotator agreement score.
To substantiate our methodology and claims, we perform extensive experimentation on multiple datasets and report the results.\footnote{An implementation of our automatic methodology is available at: \href{https://github.com/itamartrainin/Tpower5Score}{https://github.com/itamartrainin/Tpower5Score}.}

\end{abstract}

%%%%%%%%%%%%%%%%%%%%%%%%%%%%%%%%%%   
\section{Introduction}

Topic Extraction plays a key role in computational text analysis, enabling the unsupervised discovery of salient information while presenting it in a summarized and structured manner.
For this reason, there are numerous methods for addressing it  \citep{blei2003latent, abdelrazek2023topic}.
Recently, alongside the rise of LLMs, solutions are shifting towards using generative models \citep[e.g.,][]{reuter2024gptopic,garg2021keyphrase,mishra2021automatic,bosselut2019comet}. 
This trend is motivated by LLMs' ability to overcome some of the drawbacks of existing methods, such as clear topic naming, contextualization, and operating without dedicated training \cite{brown2020language}. 

Still, to validate such a use, LLMs must be assessed carefully.
Unfortunately, collecting manually annotated topic sets for direct assessment is impractical \citep{chang2009reading} due to annotators' bias \citep{reidsma2008exploiting,wich2020investigating}, and the cognitive burden of processing large amounts of information at once \citep{hoyle2021automated,nugroho2020survey}. 
For this reason, existing evaluation procedures rely on indirect approximations \citep[e.g.,][]{chang2009reading, bhatia2018topic}, rather than direct assessments of quality. Moreover, these practices result in poor Inter-Annotator Agreement (IAA) \cite{stammbach2023re}.
In the absence of standard evaluation practices, scientific works that employ topic sets often do not include a thorough evaluation of the ``correctness'' of the extracted topics \cite{livermore2017supreme,friese2018carrying,keydar2024discursive}. While these practices are common, stipulating the correctness of the topic sets without evidence is unwarranted, and equally in the case of LLM-generated topic sets.

\begin{figure}
    \centering
    \includegraphics[width=\linewidth]{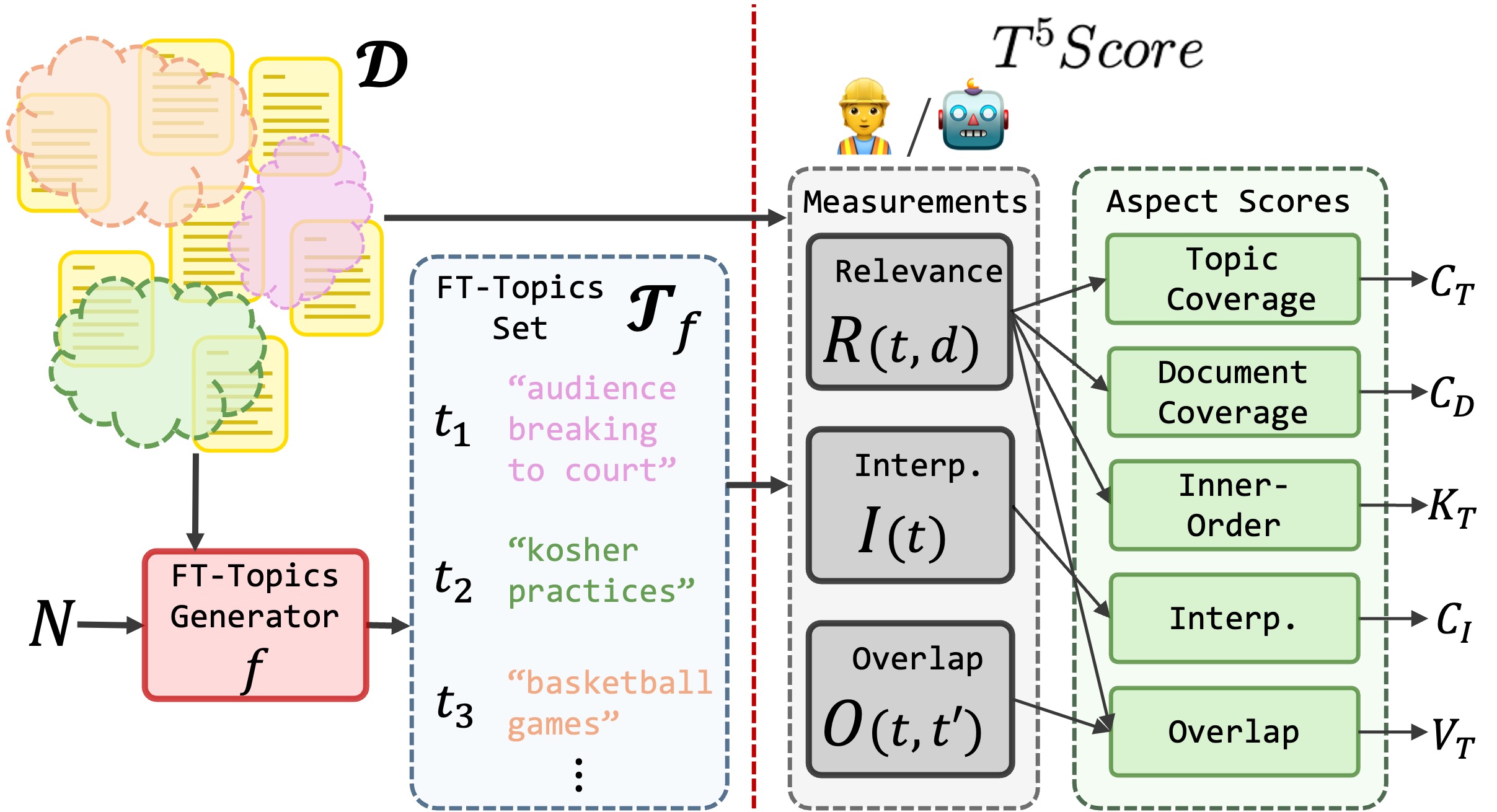}
    \caption{
        \methodname{} pipeline: A set of documents $\mathcal{D}$ encapsulating a set of shared topics is used to generate an \llmbased-topics set ($\mathcal{T}_f$) of size $N$ using a generator $f$.
        The \textit{Relevance}, \textit{Interpretability}, and \textit{(non-)Overlap} measurements are annotated by either a human or a machine.
        The resulting annotations are then used to compute the aspect-based scores.
    }
    \label{fig:system}
    \vspace{-3mm}
\end{figure}

In this paper, we develop a methodology for evaluating LLM-based Topic Sets extracted from a collection of documents. 
While the term Topic Extraction (or Topic Modeling) commonly refers to methods like LDA \citep{blei2003latent}, which output topics as word distributions, in this work, we focus on LLM-based topics that are free-text in nature and therefore are better portrayed as theme-describing titles (e.g., ``Experiences of Discrimination'').
For clarity, we will distinguish between the two cases by referring to them as \textit{Word-Distribution (\ldalike) Topics} and \textit{Free-Text (\llmbased) Topics} respectively, formally defined in \S\Ref{sec:defs}.

To overcome the challenges in direct assessment of topics, we introduce \methodname, a dedicated evaluation methodology for indirectly assessing \llmbased-topic sets, overcoming the annotator agreement barrier. 
Acknowledging the drawbacks of using aggregate metrics \citep{burnell2023rethink, kasai2021transparent}, we propose to decompose the quality of an \llmbased-topics set into three simple annotation tasks defining five scores along different aspects of quality.
See Fig. \ref{fig:system} for a schematized view of the methodology.

The remainder of this paper is structured as follows. 
First, we motivate the formulation of \methodname\ and formally define it.
Next, we conduct a human-oriented case study showing that using the methodology for manual evaluation results in high IAA scores. 
Then, we claim that the slow and expensive manual process can be automated via LLM-based labeling and support this by reporting a strong correlation with human annotators.
Finally, we show that the methodology validly reflects the expected ordering between generation systems by applying it to \llmbased-topic sets generated by different systems.
Throughout this work, alongside others, we employ a novel dataset of Holocaust survivor testimonies collected by USC Shoah Foundation (SF),\footnote{https://sfi.usc.edu/} providing collections of common yet unique experiences ideal for our work.\footnote{
    This study was established as part of an ongoing effort to study Holocaust survivor testimonies with computational tools. 
    Given the imminent passing of the last generation of Holocaust survivors, it is increasingly important that the testimonies they left be made accessible to Holocaust researchers and the public. 
    However, due to the enormity of the collected databases (tens of thousands of testimonies), only a few of them are directly read and studied. 
    Our investigation will support the development of stronger systems for processing these databases and provide a more faithful view of their trends.
}
%%%%%%%%%%%%%%%%%%%%%%%%%%%%%%%%%%%%%%%%%%%%%%%%

\section{Related Work} \label{sec:related_work}
 
\subsection{Evaluation of \ldalike-Topic Sets}

Overall, evaluation methods for topic sets can be categorized into \textit{extrinsic} and \textit{intrinsic}. 
Extrinsic methods are valuable for assessing an output used as an intermediate step in a larger system \citep{suzuki2014detection,wu2024survey,penta2022enhance}. 
However, they provide limited insight into the inherent quality of the output itself. 
Intrinsic methods, such as \citet{mimno2011optimizing}, often exhibit a weak correlation with human judgment \cite{stammbach2023re}. 
One such commonly used method utilizes the \textit{intrusion} metric \citep{chang2009reading,bhatia2018topic}, which assesses the ``coherence'' of a topic. 
However, these methodologies are exclusively designed for the evaluation of \ldalike-topic set. 

\subsection{Evaluation of \llmbased-Topic Sets}

In the case of \llmbased-topic sets, the only existing evaluation procedure relies on free-text evaluation, which most commonly assumes an available annotated data source for comparison.
Traditionally, metrics like BLEU \citep{papineni2002bleu} and ROUGE \citep{lin2004rouge} are used. 
While convenient, these metrics primarily focus on surface-level similarities, often overlooking important semantic nuances, hindering the ability to truly capture the quality of the abstraction.

Newer metrics like BERTScore \cite{zhang2019bertscore} attempt to address this by leveraging Language Models \citep{devlin2018bert,devlinetal2019bert} to gauge semantic similarity. 
While such methods offer some improvement over N-gram overlap, their performance can still be hampered in scenarios where context is lacking, like \llmbased-topics. 
In addition, semantic similarity does not capture all aspects of interest, such as whether the topics themselves are interpretable. % or the effective size of the set.
Nonetheless, the biggest hurdle for such methods is the collection and annotation process, making such data scarce.  

In \cite{lior2024leveraging}, the intrusion task is adopted to evaluate \llmbased-topic sets by treating the whole set of \llmbased-topics as a single cluster. However, this approach lacks the direct grounding put forward in this work. Additionally, while the intrusion task may be useful for assessing a single, topically coherent cluster of words, it becomes less relevant in the context of FT-Topics sets, where unrelated topics may naturally coexist.

\subsection{LM as a Judge}

Another related line of work includes the recently introduced ``Judge'' models, which serve as automatic evaluators. 
This approach attempts to leverage the strength of large models for automatically assessing the output. 
Previously, the evaluation process relied on custom models specifically trained for each use-case \citep[e.g.,][]{bhatia2018topic,gupta2014quality,peyrard2017learning}. 
However, training such models is difficult. Recognizing zero-shot and few-shot learning capabilities of LLMs \citep{brown2020language} inspired some works \citep[e.g.,][]{fu2023gptscore,huang2023chatgpt,lai2023multidimensional,kocmi2023large,wang2023chatgpt} to use LLMs as evaluators.

%Evaluating the correctness of a solution to a problem is sometimes as difficult as solving the problem itself. 
%In our work, we show that reducing the evaluation to smaller measurements simplifies it. However, further research is needed to better understand the trade-offs in such a simplification.

\subsection{Manual Evaluation of Topic Sets}

The aforementioned methodologies, including our methodology, are often designed to allow either a human or machine to perform the annotation \cite[e.g.,][]{chang2009reading,lau2014machine,nugroho2020survey}. 
Naturally, human evaluation is frequently favored \cite[e.g.,][]{chang2009reading,lior2024leveraging} due to its flexibility. 
However, manual evaluation procedures are often extremely costly and slow and therefore can only be done at a limited scale.
More importantly, these practices result in low IAA.

%%%%%%%%%%%%%%%%%%%%%%%%%%%%%%%%%%%%%%%%%%%%%%%%

\section{The {\it \methodname} Methodology} \label{sec:methodology}

\methodname\ is an indirect evaluation methodology for scoring a set of extracted topics concerning a set of queried documents.

\subsection{Formal Setting \& Definitions}\label{sec:defs}

The literature often associates the term Topic Extraction (or Topic Modeling) with LDA-like methods that output topics as word distributions (henceforth: \textit{Word-Distribution (\ldalike) Topics}). 
In this work, we focus on sets of topics that are free text in nature. 
That is, sets of textual descriptions inferring topics in the documents (henceforth: \textit{Free-Text (\llmbased) Topics}) formally defined as a list of \texttt{strings} and are denoted by $\{t\in\mathcal{T}_{f}\}$. 
For example, an \llmbased-topics set may include the topics ``Transportation to Concentration Camps'',
capturing a major theme in Holocaust survivor testimonies (see Table \ref{tab:validation_examples} for more examples). 

We denote a system that generates \llmbased-topics sets with $f(N, \mathcal{D})$. 
Such a system receives a parameter $N$, the number of expected topics, and a set of documents $\{d\in\mathcal{D}\}$, where $M=|\mathcal{D}|$. 

\methodname\ assesses the quality of an \llmbased-topics set by performing 3 annotation tasks (henceforth, \textit{measurements}): $I(t)$, $R(t,d)$, and $O(t,t')$. 
A \textit{measurement} is defined as a function of the topic set and the sample and is directly annotated by either a human or a machine (refer to \S\ref{sec:Defining_the_Quality_of_a_Title_Set} for definitions). 
In this work, we define the explicit functionality of the measurement as annotation guidelines or prompts (refer to Appendix \S\ref{app:annotation_guidelines}, \S\ref{app:llm_prompts}). 

The measurements are then used to formulate five scores representing different aspects of quality: $C_T$, $C_D$, $K_T$, $C_I$, $V_T\in\left[0,1\right]$ (see \S\ref{sec:Defining_the_Quality_of_a_Title_Set} for formal definitions). See Fig. \ref{fig:system} for a schematized view of the methodology.

\subsection{Defining the Quality of an \llmbased-topics Set} \label{sec:Defining_the_Quality_of_a_Title_Set}

We say that an \llmbased-topics set is a "good" set concerning a collection of reference documents if it scores high on the following five aspects of quality:

\subsection*{Aspect 1: Interpretability} 

The themes that the topics in the set inflict may not always be clear to a user. For example, in the context of Holocaust experiences, it is hard to decipher the theme induced by the topic ``\textit{sadness}''.
The range of emotions present during such an experience makes it difficult to understand what specific aspect of the experience the topic is meant to highlight. 

We define the aspect of Interpretability to assess whether the topics in the set correspond to themes in the corpus.
A topic describes a theme if it is interpreted as that theme by the annotators. Formally, we define:

\begin{equation}
    C_I = \frac{1}{N}\sum_{t\in\mathcal{T}_{f}}I(t)
\end{equation}

Where $I(t)$ denotes the interpretability measurement that accepts a topic and outputs a score in $[0,1]$ for the degree to which the annotator can infer the theme induced by the topic.

\subsection*{Aspect 2: Topic-Coverage}

A ``good'' set of topics is a set that covers (``\textit{tell a story of}'')  the queried documents. 
Specifically, this aspect quantifies the extent to which the themes induced by the topics in the set capture the \textit{major themes} in the corpus.
A major theme is defined as a theme that recurs broadly across the corpus.
Hence, a topic that is relevant to many documents in the corpus (i.e., covers the corpus) is a topic capturing a major theme.
For example, within experiences of deportation, \textit{``Transportation to Concentration Camps''} is a major theme since it is likely to cover most, if not all, deportation experiences. 
Formally, we define: 

\begin{equation}
    C_T = \frac{1}{N}\frac{1}{M}\sum_{t\in\mathcal{T}_{f}, d\in\mathcal{D}}R(t,d)
\end{equation}

Where $R(t,d)$ denotes the relevance measurement. 
For each topic-document pair in $\mathcal{T}\times\mathcal{D}$, the measurement returns a score in $[0,1]$ expressing the relevance of the topic to the document, evaluating the description in context.

\subsection*{Aspect 3: Document Coverage}

Another desired quality of a topic set is that no major theme was omitted.
To gauge this notion, we set a lower bound on the true coverage by measuring the coverage level of the least-covered document. 
Low Document Coverage scores indicate that better coverage could be achieved by adding additional topics to the set.
Formally, we define,

\begin{equation}
    C_D = \min_{d\in\mathcal{D}}\max_{t\in\mathcal{T}_{f}}\{R(t,d)\}
\end{equation}

To mitigate the influence of noisy out-of-distribution documents in the reference set, we have experimented with both Log-Sum-Exponent \cite{nielsen2016guaranteed} and p-Norm function as smooth approximations of the minimum. 
We found that the relative ordering of different generation systems remains the same for all implementations and therefore choose to use the simple $\min(\cdot)$ function. 

\subsection*{Aspect 4: (non-)Overlap}

One of the main challenges with topic sets is that topics in the set often describe themes that are abstractions of one another or are the same de facto. 
A well-constructed topic set aims to minimize such redundancies. 
Ergo, this aspect quantifies the extent to which the themes induced by topics in the set overlap.
For example, the descriptions \textit{``Transportation to Concentration Camps''} and \textit{``Transportation by a Wagon''} may refer to the same theme. 
Formally: 

\begin{equation} \label{eq:non_overlap}
V_T = \frac{1}{N}\sum_{t\in\mathcal{T}_{f}} \left[1 - \max(v_{\text{def}}(t), v_{\text{cov}}(t)) \right] 
\end{equation}

\begin{equation} \label{eq:def_overlap}
v_{\text{def}}(t)=\max_{t'\in\mathcal{T}_{f}, t \neq t'}{O(t,t')}
\end{equation}

\begin{equation} \label{eq:cov_overlap}
v_{\text{cov}}(t)=\max_{t'\in\mathcal{T}_{f}, t \neq t'}{\sum_{d\in\mathcal{D}}R(t,d)\cdot R(t',d)}
\end{equation}

Where $O(t_1,t_2)$ measures the overlap in the definition. It receives a pair of topics and outputs a score in $[0,1]$ for the degree to which themes expressed by the two topics overlap.

Intuitively, Eq. \ref{eq:def_overlap} captures the overlap in the definition of two given topics, reflected by the annotator’s understanding of the theme induced by the two topics. 
Alongside, Eq. \ref{eq:cov_overlap} captures the overlap in coverage, that is, if the two topics cover the same documents, they may represent the same themes.
Thus, Eq. \ref{eq:non_overlap} captures the average non-overlap between the topics in the set.

\subsection*{Aspect 5: Inner-Order}

Assesses whether the topics in the set are ordered by their importance.
In some cases, although not all, the order of topics reflects importance, where more important topics precede less important ones in the set. 
For example, \textit{``Transportation to Concentration Camps''} should be ordered before \textit{``Transportation by a Wagon''} as the latter represents a less major theme concerning the first. 

If the topic set is well-ordered, its inner order should reflect the order of the topic's importance. 
Formally,

\begin{equation}
K_T = \max(0,\tau(\mathcal{T}_{f}, \mathcal{T'}))
\end{equation}

where $\tau(\cdot)$ is the Kendall $\tau$ ranking correlation coefficient \cite{kendall1948rank}, and $\mathcal{T'}$ is a re-ordering of $\mathcal{T}_{f}$ according to the mean relevance:

\begin{equation}
r_t = \frac{1}{M}\sum_{d\in\mathcal{D}}R(t,d)
\end{equation}

Not all systems reflect an inner-order however, by including this score, we hope to motivate the generation of sets that do reflect it.

\subsection{Aggregate Score}

While scoring a topic set using individual components is beneficial, in practice, an aggregate score is often preferred. Thus, in Appendix \ref{app:aggregate_score} we present a simple aggregation measure and use it to score the different systems presented in this paper. Nevertheless, as the relative importance of the different components varies considerably between use cases, we advise caution in deciding which aggregation function to use. We defer a deeper discussion of this topic to future work.

%%%%%%%%%%%%%%%%%%%%%%%%%%%%%%%%%%%%%%%%%%%%%%%%

\section{Datasets}\label{sec:data}

\subsection{USC-SF Survivor Testimonies}
A collection of Holocaust survivor testimonies put together by USC-SF.\footnote{Recorded and transcribed testimonies are available upon request to USC Shoah Foundation.}
This collection comprises stories recounted by survivors based on their unique experiences and perspectives during the Holocaust. 
Each testimony naturally describes different experiences, but many of the themes do recur, albeit in a variety of circumstances, times, and places. 
Therefore, this dataset offers sub-collections of documents on relatively constrained domains, making it ideal for this work. We are further motivated by the use of this dataset in recent computational modeling work \citep{wagner2022topical,wagner2023event,ifergan2024identifying,shizgal2024computational}.

The testimonies (see examples in Table \ref{tab:seg_examples}) were collected as part of an oral interview in English between a survivor and an interviewer. 
The recordings were later transcribed into text. 
Since the testimony is told as part of an interview, the data is segmented according to the speaker sides, where most of the time survivors share their experiences while the interviewer guides the testimony with questions.
Testimony lengths range from 2609 to 88105 words, with a mean length of 23536 words \citep{wagner2022topical}.

We use an existing labeling of the dataset performed by SF (henceforth: \textit{Domain-Names}), which identifies testimony segments that are related across survivors. 
The domain-names are based on a pre-defined human-generated hierarchical ontology where segments of roughly 1 minute (of audio time) were labeled with one or more human-written ontology classes. 
For our purposes, we have clustered segments from multiple testimonies that share a domain-name, to form \textit{domains}. 
These domains represent common experiences with shared themes that could be used in our experiments.
A single testimony may contain multiple non-consecutive segments sharing a domain-name.
For this reason, we define a document as a concatenation of all segments in a single testimony sharing a domain-name.
In this work, we used a subset of 21 domains that describe relatively constrained experiences.
See Table \ref{tab:domains-dist} in the Appendix for data distributions and domain-names.
 
\subsection{Multi-News}
We utilized the Multi-News dataset \citep{fabbri2019multi}. 
This dataset contains event-centered clusters of news reports (i.e. \textit{domains}) published by different agencies. 
The reports in the dataset are of average length of 260 words.
Throughout our experiments, we used a subset of 71 domains, each consisting of at least 6 and no more than 10 reports.

%%%%%%%%%%%%%%%%%%%%%%%%%%%%%%%%%%%%%%%%%%%%%%%%

\begin{table}
    \centering
    \begin{tabular}{lccc} 
    \toprule
    Measurement & \# Items & \# Anno. & Agreement \\  
    \midrule
    Interp. & 550  & 3 & 0.66 \\
    Relevance & 1583  & 4 & 0.67 \\
    Overlap & 464  & 2 & 0.78 \\
    \bottomrule
    \end{tabular}
    \caption{
        Agreement scores achieved on each annotation task, including the number of items tagged, the number of annotator participants, and the resulting Krippendorff-$\alpha$ score.
        All annotators tagged all items.
        The definition of an ``item'' may vary across tasks, refer to \S\ref{sec:Defining_the_Quality_of_a_Title_Set} for measurement definitions.
    }
    \label{tab:int_ann_agg}
    \vspace{-3mm}
\end{table}

\section{IAA Study} \label{sec:a_human_centric_evaluation}

In the following, we demonstrate that using \methodname\ in evaluation procedures performed manually by human annotators results in high inter-annotator agreement (IAA). We conduct a human-oriented case study designed according to the methodology and report the observed IAA score.

\subsection{Experimental Setup}
We recruited 4 fluent English-speaking in-house annotators.
The participants were asked to perform the interpretability ($I(t)$), relevance ($R(t,d)$), and overlap ($O(t,t')$) measurements according to the guidelines defined in \S\ref{sec:Defining_the_Quality_of_a_Title_Set} and explicitly described in Appendix \ref{app:annotation_guidelines}. 
The annotations were performed over \llmbased-topics sets generated from the USC-SF dataset using GPT3.5 (see Appendix \ref{app:llm_prompts} for the generation prompt). 
During each session, the annotators were introduced to a set of topics and $M=10$ in-domain reference documents, maintaining full item overlap between annotators. Overall, each annotator was introduced to a total of 21 topic sets and 210 reference documents, see final item counts in Table \ref{tab:int_ann_agg}.

The annotators received guidance both in-person and through written annotation guidelines. 
Importantly, the annotators were asked to make no assumptions based on previous knowledge that did not appear in the context of the reference documents. 
During the annotation process, we followed the conclusions from \citet{graham2013continuous} and used Continuous Scale Rating on the scale of $[0-100]$. 

\begin{table*}[t]
\centering
\begin{tabular}{lc?ccc} 
\toprule
\textbf{Model} & \textbf{Quantization} & \textbf{Relevance} & \textbf{Overlap} & \textbf{Interpretability} \\  
\midrule
GPT 4 & - & \textbf{0.66} & 0.86 & 0.63 \\
GPT 3.5 & - & 0.50 & 0.79 & \textbf{0.73} \\
GPT 4o Mini & - & 0.61 & \textbf{0.87} & 0.61 \\
\midrule
LLaMA 3 (8B) & None & 0.45 & 0.85 & 0.54 \\
LLaMA 3 (70B) & 4-bit & 0.48 & 0.24 & 0.29 \\
LLaMA 3 (70B) & None & \underline{0.62} & \textbf{\underline{0.87}} & \underline{0.66} \\
\midrule
Mixtral (8x7B) & 4-bit & 0.43 & 0.83 & 0.65 \\
Mixtral (8x7B) & None & 0.50 & 0.73 & 0.65 \\
\bottomrule
\end{tabular}
\caption{Spearman correlation between LLM and mean human annotations. The best overall model for each measurement is boldfaced, and the best open-source alternative is underlined.}
\label{tab:judge}
\end{table*}

We note that contemporary LLMs rarely output uninterpretable content. Therefore, for a reliable study of the interpretability measurement, we synthetically increased the number of uninterpretable \llmbased-topics (negative items).
Since simple topic corruption, like invalid words or phrases, can be easily distinguished, we were looking for semantically coherent descriptions that are not indicative of any theme in the corpus. 
We opted to use GPT-4 to corrupt valid topics.
Examples and generation prompts can be found in Appendix \ref{app:llm_prompts}, Table \ref{tab:title_corr}.

\subsection{Results}

Table \ref{tab:int_ann_agg} reports the resulting IAA scores (Krippendorff-$\alpha$ \cite{krippendorff2011computing}), indicating high levels of agreement between the different measurements, overcoming the agreement barrier and highlighting the effectiveness of \methodname\ for manual evaluation procedures. 

%%%%%%%%%%%%%%%%%%%%%%%%%%%%%%%%%%%%%%%%%%%%%%%%
\section{LLMs as Automatic Evaluators} \label{sec:LLMs_as_Automatic_Evaluators}

\methodname\ decomposes the evaluation process into small and simple annotation tasks. 
This framing enables the use of off-the-shelf LLMs as autonomous annotators, making this methodology applicable independently of slow and expensive human annotation procedures. 
We experiment with different popular LLMs, reporting comparable performance to human annotators.   

\subsection{Experimental Setup}

We employ popular LLMs, including OpenAI-GPT \citep{achiam2023gpt, brown2020language}, Mixtral \cite{jiang2024mixtral} and Meta-LLaMA (see model versions in Appendix \ref{app:llm_versions}) as predictors on the different measurement tasks ($I$, $R$, $O$). We then compared the predictions to the human labels collected in \S\ref{sec:a_human_centric_evaluation}.
For the last two, we used both no-quantization and 4-bit quantization setting. 
For each measurement, a prompt was written (see Appendix \ref{app:llm_prompts}) for querying the model based on the measurement definitions in \S\ref{sec:Defining_the_Quality_of_a_Title_Set}. 

\subsection{Results}

Table \ref{tab:judge} reports the Spearman correlation \cite{spearman1961proof} between the LLM's predictions and the mean human score. 
The results indicate that LLMs can simulate human annotations, achieving a high overall correlation to the human baseline.
Although the best model varies in each measurement, we note the GPT-4's dominance, along with LLaMA-3 (70B) without quantization as a reasonable open-sourced alternative.
To further substantiate this claim, we include additional results in Appendix \ref{app:additional_results}.
This includes other correlation measures in Table \ref{tab:judge_full}, showing the same conclusions. 
We also report correlations between each annotator and the mean human score; see Table \ref{tab:human_to_mean}.
The high correlation further stresses the reliability of our conclusions.

%%%%%%%%%%%%%%%%%%%%%%%%%%%%%%%%%%%%%%%%%%%%%%%%

\begin{figure*}
    \centering
    \includegraphics[width=\linewidth]{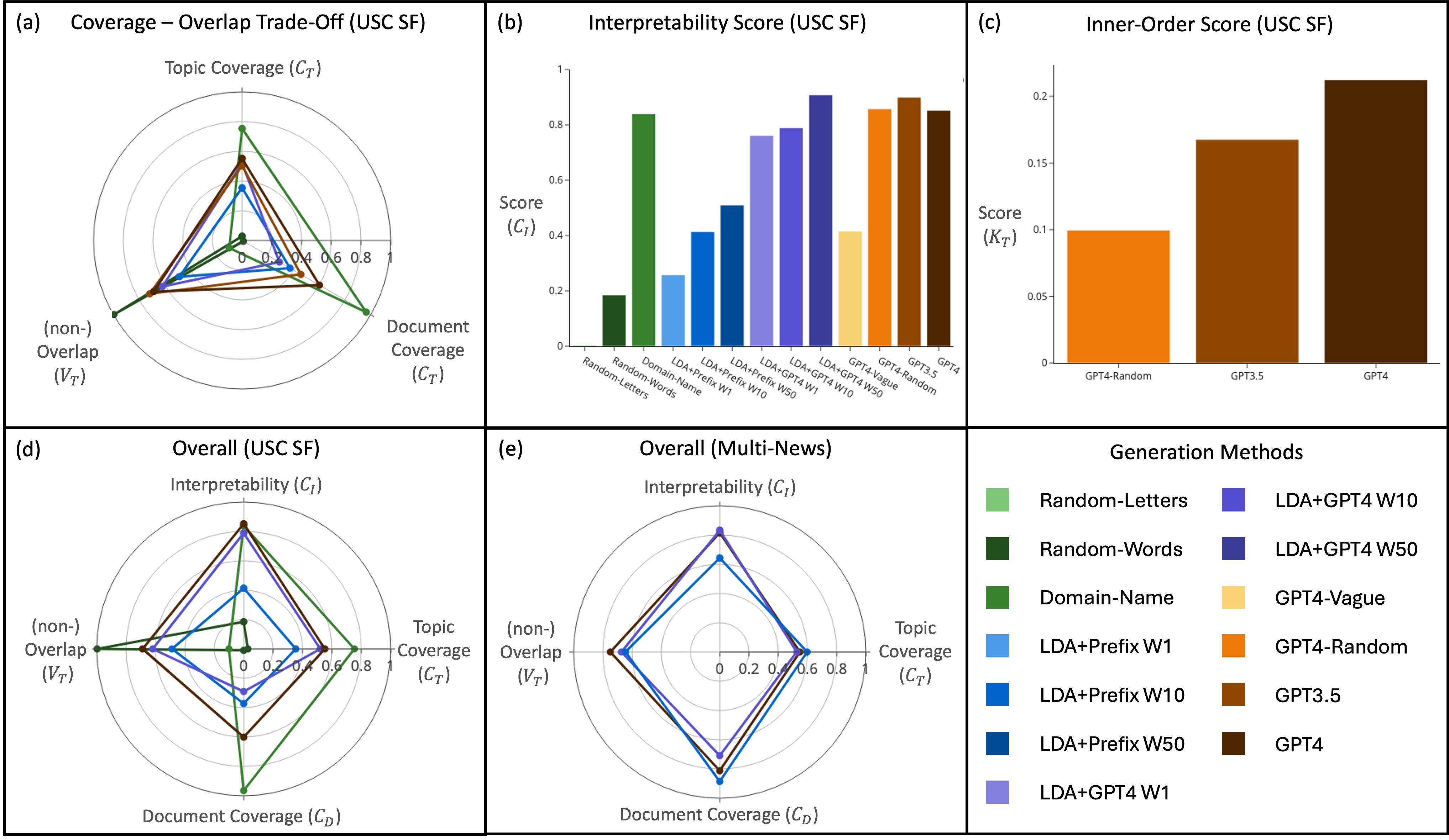}
    \caption{Applying \methodname\ on generated \llmbased-Topic Sets; (a) Coverage (Topic Cov. and Doc. Cov.) and the non-Overlap trade-off; (b) Interpretability scores; (c) Inner-Order scores achieved by LLM-based systems; (d) overall comparison of representative systems; (e) overall comparison for the Multi-News dataset.}
    \label{fig:validation_aspects}
    \vspace{-3mm}
\end{figure*}

\section{Applying \methodname\ on \llmbased-Topic Sets} \label{sec:case_study}

Thus far, we motivated the design of \methodname\ (\S\ref{sec:methodology}) and proposed an automatic implementation achieving human-like performance (\S\ref{sec:LLMs_as_Automatic_Evaluators}). 
Yet, it remains to show whether this framing is valid, namely whether it reflects the expected ordering between different systems. 

Often, such a study is performed by comparing the resulting relative order of systems to an ordering based on human preferences \citep[e.g.,][]{papineni2002bleu}. 
However, since human annotation of \llmbased-topic sets is unreliable (\S\ref{sec:related_work}), this option is unfeasible. 
Indeed, such analysis is often omitted from other topic set evaluation works \citep[e.g.,][]{chang2009reading,bhatia2018topic}, where only the ability to perform the proxy task (e.g., recognizing an intruder in intrusion tests) is compared to humans (as in  \S\ref{sec:LLMs_as_Automatic_Evaluators}). 

Instead, we devise case studies to empirically support the validity of our methodology. 
We carefully design generation systems where the relative ordering between at least a subset of them is clear and show that this ordering is reflected by \methodname. 
Since it could be argued that some of these systems are too simplistic, we also synthetically compile a set of human-written topics and study \methodname's performance on it too.

\subsection{Generation Systems} \label{sec:generation_systems}
For this study, we implement the following \llmbased-topics generation systems. Refer to Appendix \ref{tab:validation_examples} for examples and Appendix \ref{app:llm_prompts} for relevant prompts. 

\paragraph{LDA-Based.} 

\begin{enumerate}[leftmargin=*,topsep=0pt,itemsep=-1ex,partopsep=1ex,parsep=1ex]
    \item \textbf{LDA+Prefix} topics are a quoted, comma-separated list of the top $k$ words of each LDA cluster. A prefix ``\textit{The theme defined by the following set of words:}'' is then prepended.
    \item \textbf{LDA+GPT4} topics are generated by prompting GPT-4 with the top $k$ words of each LDA cluster. 
\end{enumerate}

\paragraph{LLM-Based.}
\begin{enumerate}[leftmargin=*,topsep=0pt,itemsep=-1ex,partopsep=1ex,parsep=1ex]
    \item \textbf{GPT} GPT is used to generate common topics from a sample of documents. 
    \item \textbf{GPT4-Random} random topics uniformly chosen from the union of \llmbased-topics sets from all samples, as generated by GPT-4.
    \item \textbf{GPT4-Vague} uses the topic corruption procedure from \S\ref{sec:data} to corrupt all of the topics generated by GPT-4.
\end{enumerate}

\paragraph{Synthetic.}
\begin{enumerate}[leftmargin=*,topsep=0pt,itemsep=-1ex,partopsep=1ex,parsep=1ex]
    \item \textbf{Random-Letters} produces topics comprised of random sequences of English letters.
    \item \textbf{Random-Words} generates topics by combining random, yet real, English words. 
    \item \textbf{Domain-Name} Topic sets are created by assigning the same human-written domain-name to every instance in the set (if applicable by the dataset).
\end{enumerate}

\subsection{Automatically Generated Sets}

We aim to study whether \methodname\ reflects the expected order between the automatic generation systems devised in \S\ref{sec:generation_systems}.
We have used each system to generate topic sets based on the USC-SF ($N=10$, $M=8$) and Multi-News ($N=3$, $M\in[6,10]$) datasets.
In addition, we employed ``\textit{gpt-4o-mini}'', for its cost-effectiveness, as the judge model for automatically running \methodname. 

By design, we expect that the Synthetic methods will score on the extreme ends of the aspect scales.
On the contrary, we expect that the more naturalistic methods will score lower on the separate aspects but achieve a higher and more balanced combined score. 
We anticipate that LLM-based models will achieve higher scores than LDA-based due to their advanced capabilities, context understanding, and human-like fluency.
Furthermore, we predict that aspect scores will rise as $k$, the number of words describing LDA-based topics, increases. 

Figures \ref{fig:validation_aspects}(a)-(e) showcase a summarized breakdown of the aspect scores achieved on the different datasets.
Please refer to Appendix \ref{app:additional_results} for full results and extended analysis.

Fig. \ref{fig:validation_aspects}(a) presents a system comparison through the Coverage and (non-)Overlap aspects. 
As shown, the relative ordering of the systems aligns with expectations. 
Furthermore, this comparison highlights a significant trade-off in \llmbased-topic set generation practices: while high-level topics enhance coverage, they also lead to increased overlap.
The figure effectively depicts this trade-off, as synthetic methods excel in one aspect at the cost of another, whereas non-synthetic methods exhibit greater variability, resulting in lower scores across individual aspects (refer to Table \ref{tab:validation_examples} in the Appendix for examples). 

Fig. \ref{fig:validation_aspects}(b) illustrates how the methodology accurately reflects the low interpretability of synthetic methods, except for Domain-Names, which are human-written and therefore score highly. 
As expected, methods incorporating LLMs achieve high scores due to their strong fluency \cite{yang2023exploring,lai2023multidimensional,jiao2023chatgpt}, while LDA-based methods score lower. 
The figure further shows that LDA topics generated with more words, hence more interpretable, achieve higher scores.

Fig. \ref{fig:validation_aspects}(c) reports the Inner-Order scores of LLM-based systems, including only those capable of reflecting ordering. 
Fig. \ref{fig:validation_aspects}(d) shows an overall comparison of the different aspects but Inner-Order. 
Finally, Fig. \ref{fig:validation_aspects}(f) presents aspect scores for the Multi-News dataset.
All show consistent trends substantiating the validity of the methodology.

\subsection{Human Generated Sets}

In this experiment, we seek to study how \methodname\ behaves when applied to human-generated topic sets. 
In the absence of ground truth, we compiled synthetic human-generated sets using USC-SF's human-written domain-names. We selected $N=5$ domain-names (\{\textit{``family interactions''}, \textit{``deportation to concentration camps''},  \textit{``transfer from concentration camps''}, and \textit{``post-liberation recovery''}\}) for representing experiences that do not occur simultaneously in time. It is therefore unlikely that one document is related to more than one topic. 

To simulate multiple such sets, we repeatedly sampled sets of reference documents from the union of the domains.
Overall, we collected 50 sets of $M=10$ reference documents, each corresponding to the set of human-generated topics.
Knowing the true relation between the document and the human-generated topic set enables direct computation of the aspect scores.
In addition, the documents were used to generate topic sets using the different automatic systems presented in \S\ref{sec:generation_systems}.

\begin{figure}
    \centering
    \includegraphics[width=\linewidth]{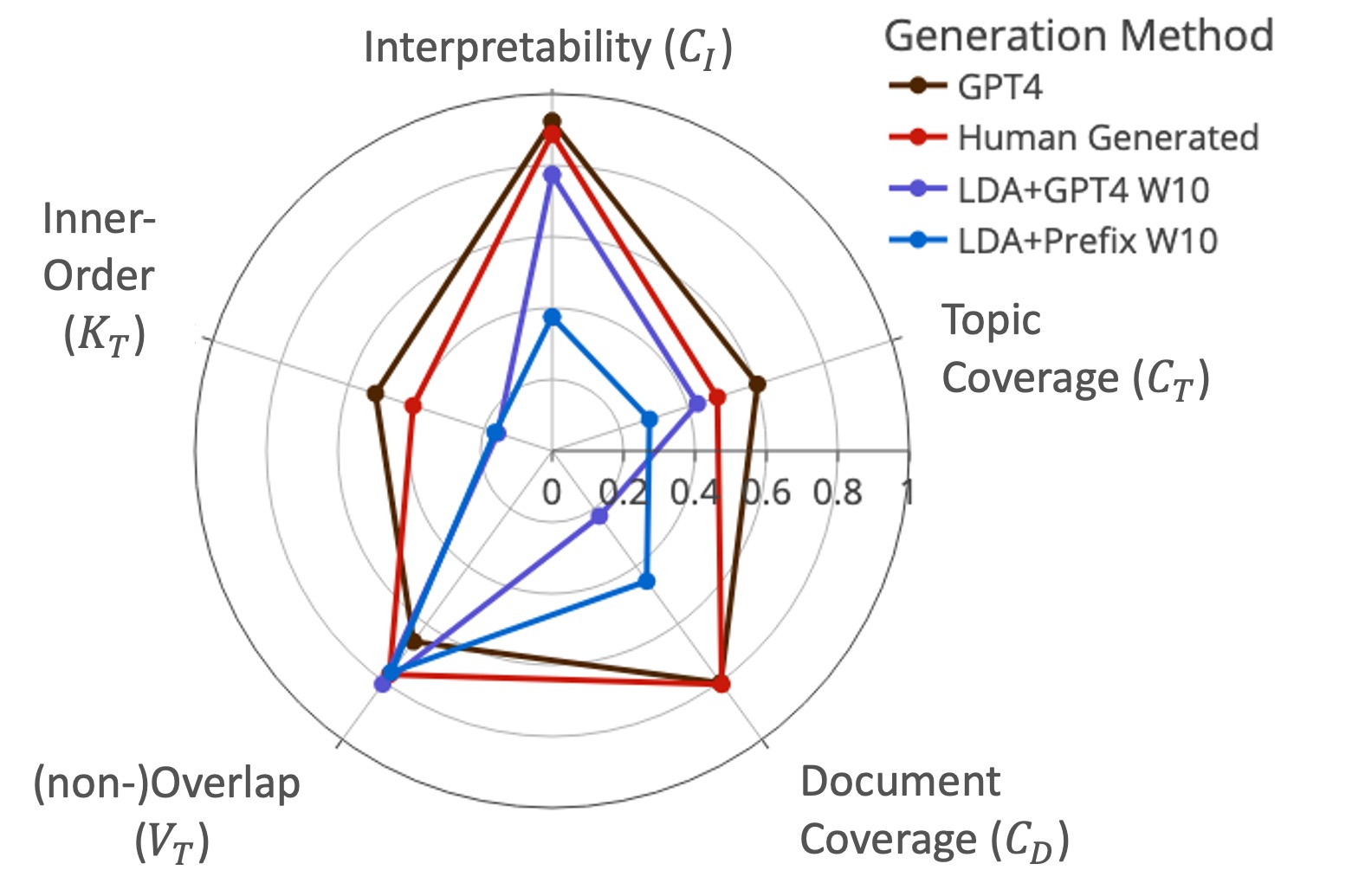}
    \caption{Human Generated Topic Sets Comparison.}
    \label{fig:human_baseline}
    \vspace{-3mm}
\end{figure}

A comparison of the results is reported in Fig. \ref{fig:human_baseline}.
The figure demonstrates that the expected ordering of systems is preserved, where the human-generated set achieves a high overall score, comparable only to GPT-4.
Specifically, in the Interpretability aspect, the human-generated set attains a high score, as expected, given that each aspect was explicitly written by humans. 
Regarding Topic Coverage, while the human-generated set performs well, it is outperformed by GPT-4, which was explicitly tailored to each document set. 
For Document Coverage, the human-generated set achieves a high score due to the broad scope of domain-names.
The (non-)Overlap score is high, aligning with the set’s design.
Lastly, in terms of Inner-Order, the human-generated set scores well but is surpassed by GPT-4 for the same reason as for the Topic Coverage aspect. 
These results further support the validity \methodname. 
See Appendix \ref{app:additional_results} for full results.

%%%%%%%%%%%%%%%%%%%%%%%%%%%%%%%%%%%%%%%%%%%%%%%%

\section{Conclusion}

We presented \methodname, an evaluation methodology that decomposes the quality of an \llmbased-topics set into aspects quantifiable by easy-to-perform measurements.
We justified using our methodology by conducting extensive experimentation on a novel dataset of Holocaust survivor testimonies as well as a dataset of news reports. 
The results showed that our methodology achieves a high level of IAA on manual evaluation procedures. 
They also demonstrated that it can be automatically performed by judge models that simulate human annotations reliably.
Furthermore, the results confirmed that our methodology validly reflects the expected relative order between the tested systems in cases where the order is known beforehand.

Given the centrality of the task and the great difficulty in evaluating it reliably, we hope that the methodology proposed here will assist in the development of demonstrably stronger topic set generation systems.

%%%%%%%%%%%%%%%%%%%%%%%%%%%%%%%%%%%%%%%%%%%%%%%%

\section*{Limitations}

Some limitations in our work are a direct result of using the USC-SF Survivor Testimonies as a data source. 
First, multiple parts of the same experience may be scattered throughout the testimony, defying the story timeline.
To handle this problem, we have defined a document as the concatenation of all of those segments. 
However, each such segment may have been told in a different context, which could influence the interpretation of the text. 
Secondly, the prior ontology labeling of the segments was done on segments of constant 1-minute length. 
This coarse segmentation may cause unrelated information to be included in the segment, as well as a misplacement of small but crucial segments.

Other limitations stem from the use of LLMs. 
First, LLMs are black box models, often trained by commercial companies that do not disclose their inner workings, limiting the replicability of the results. 
Second, these models are extremely expensive to use, either as services or by running them locally on multiple high-end GPUs. 
Since our method requires employing such models, the high cost may pose a limitation in some contexts. However, we expect this cost to rapidly decline in the near future.

%%%%%%%%%%%%%%%%%%%%%%%%%%%%%%%%%%%%%%%%%%%%%%%%

\section*{Ethics Statement}

Annotation in this project was done by in-house annotators, who were employed by the university and given instructions and explanations about the task beforehand. 
During the in-person presentation of the task, the annotators were informed of the sensitive nature of the data. 
The annotators were allowed to skip sections that may affect their well-being and were asked to report such cases. 
In addition, the annotators were invited to discuss any discomfort with the moderator. 

As for the testimonies, we abided by the instructions provided by the SF. 
We note that the witnesses identified themselves by name, and so the testimonies are open and not anonymous by design. 
We intend to release our scripts, but those will not include any of the data received from the archive; the data and trained models used in this work will not be given to a third party without the consent of the relevant archives. 
The testimonies can be accessed for browsing and research by requesting permission from the relevant archive. 
Some of them are openly available online through designated websites. 

Holocaust testimonies are, by nature, sensitive material. 
Users should exercise caution when applying LLMs for Holocaust testimonies to avoid incorrect representation of the told stories.

\bibliography{anthology,custom}
\bibliographystyle{acl_natbib}

\onecolumn

%%%%%%%%%%%%%%%%%%%%%%%%%%%%%%%%%%%%%%%%%%%%%%%%

\appendix

%%%%%%%%%%%%%%%%%%%%%%%%%%%%%%%%%%%%%%%%%%%%%%%%

\section{USC-SF Data Distributions} \label{app:data}

\begin{figure}
    \centering
    \includegraphics[width=0.5\linewidth]{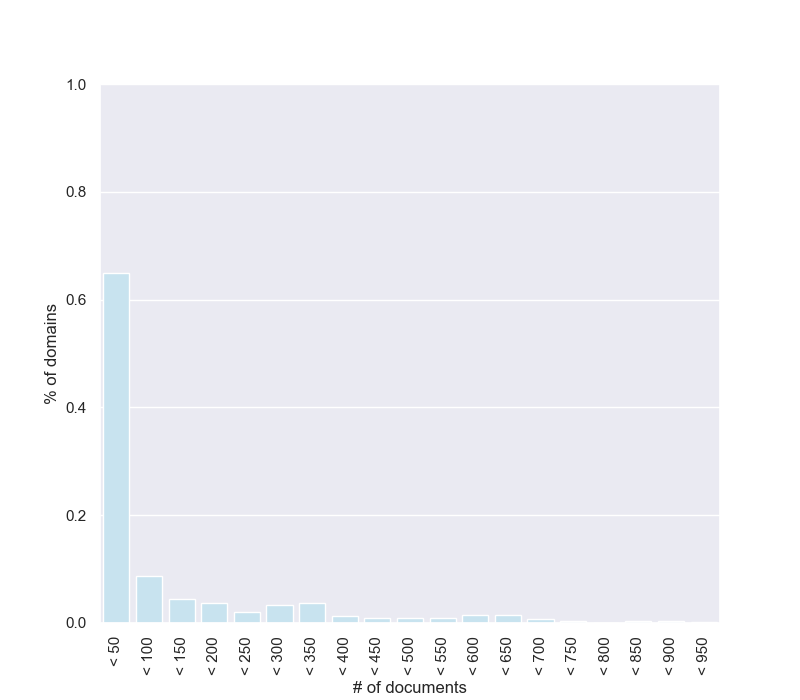}
    \caption{Size distribution of domains in terms of number of documents. Note that most domains contain less than 50 documents.}
    \label{fig:document_counts}
\end{figure}

Overall, in the data processing stage of the Holocaust survivor testimonies, we have extracted 572 different sets of documents, i.e., domains, 
each attributed using a unique domain-name. The sets where of size $M\in[1, 999]$ where $M_{ave}=105$ and average document length of 86 sentences. For our purposes, we have selected a subset of 21 domains. Table \ref{tab:domains-dist} depicts the domain names of these sets named by the USC-SF, the number of documents in the set, and the mean length of a document in the domain. Fig. \ref{fig:document_counts} shows an overall distribution of all the available domains. Table \ref{fig:overall-all} includes examples of testimony segments and their corresponding ontology labels assigned by USC-SF.

\begin{table}[h]
\centering
\begin{tabular}{lcc} 
\toprule
\textbf{Experience} & \textbf{\# Documents} & \textbf{Ave. length (sentences)} \\ [0.5ex] 
\midrule
Deportation To Concentration Camps & 308 & 41.5 \\
Family Interactions & 900 & 124.9 \\
Living Conditions & 815 & 101.1 \\
Forced Marches & 345 & 51.6 \\
Jewish Religious Observances & 700 & 83.7 \\
Anti-Jewish Regulations & 597 & 49.9 \\
Antisemitisem & 672 & 55.0 \\
Armed Forces & 541 & 70.5 \\
Food and Drink & 449 & 61.9  \\
Forced Labor & 530 & 162.8 \\
Hiding & 450 & 118.7  \\
Housing Conditions & 356 & 57.3 \\
Immigration & 633 & 113.2  \\
Jewish Holidays & 503 & 62.2 \\
Kapos & 138 & 64.4  \\
Liberation & 567 & 36.3 \\
Military Activities & 551 & 71.3  \\
Post-Liberation Recovery & 398 & 42.6 \\
Sanitary and Hygienic Conditions & 178 & 39.4 \\
Soldiers & 621 & 64.6 \\
Transportation Routes & 347 & 40.8 \\
\bottomrule
\end{tabular}
\caption{Domain data distributions. Each domain is labeled by USC's annotators. Each document is a concatenation of all segments in a testimony that were labeled as belonging to this experience.}
\label{tab:domains-dist}
\end{table}

%%%%%%%%%%%%%%%%%%%%%%%%%%%%%%%%%%%%%%%%%%%%%%%%

\section{LLM Prompts} \label{app:llm_prompts}
Throughout our work, we have used the following prompts when employing LLMs:
\subsection*{Relevance Score}\par
\underline{\textit{System Prompt}}:
\begin{displayquote}
{\fontfamily{qcr}\selectfont
You are a helpful Holocaust researcher assistant. You will perform the following instructions as best as you can.\\
You will be presented with a topic and a text. Rate on a scale of 1 to \{\textit{max-rate}\} whether the topic describes a part of the text (``1'' = does not describe, ``\{\textit{mid-rate}\}'' = somewhat describes, ``\{\textit{max-rate}\}'' = describes well). \\
Provide reasoning for the rate in one sentence only. \\\\
Please output the response in the following JSON format:\\
\{\\
    ``rate'': \text{<rate>}\\
    ``reasoning'': \text{<reasoning>}\\
\}
}
\end{displayquote}\par
\underline{\textit{User Prompt}}:
\begin{displayquote}
{\fontfamily{qcr}\selectfont
    Topic: ``\{\textit{topic}\}'',\\
    Text: ``````\{\textit{document}\}''''''
}
\end{displayquote}

\subsection*{non-Overlap Score}\par
\underline{\textit{System Prompt}}:
\begin{displayquote}
{\fontfamily{qcr}\selectfont
You are a helpful Holocaust researcher assistant. You will perform the following instructions as best as you can. 
You will be presented with two topics: topic1 and topic2. Rate on a scale of 1 to \{max-rate\} whether topic1 have the same meaning as topic2 (``0'' = different meaning, ``\{mid-rate\}'' = somewhat similar meaning, ``\{max-rate\}'' = same meaning).
Provide reasoning for the rate in one sentence only. 

Please output the response in the following JSON format:\\
\{\\
    ``rate'': \text{<rate>}\\
    ``reasoning'': \text{<reasoning>}\\
\}
}
\end{displayquote}\par
\underline{\textit{User Prompt}}:
\begin{displayquote}
{\fontfamily{qcr}\selectfont
    topic1: ``\{\textit{topic1}\}'',\\
    topic2: ``\{\textit{topic2}\}''
}
\end{displayquote}

\subsection*{Interpretability Score}\par
\underline{\textit{System Prompt}}:
\begin{displayquote}
{\fontfamily{qcr}\selectfont
You are a helpful Holocaust researcher assistant. You will perform the following instructions as best as you can. 
You will be presented with a title representing a topic. Rate on a scale of 1 to \{max-rate\} whether the topic represented by the title is interpretable to humans (``0'' = not interpretable, ``\{mid-rate\}'' = somewhat interpretable, ``\{mid-rate\}'' = easily interpretable).
Provide reasoning for the rate in one sentence only.

Please output the response in the following JSON format:\\
\{\\
    ``rate'': \text{<rate>}\\
    ``reasoning'': \text{<reasoning>}\\
\}
}
\end{displayquote}\par
\underline{\textit{User Prompt}}:
\begin{displayquote}
{\fontfamily{qcr}\selectfont
    topic1: ``\{\textit{topic1}\}'',\\
    topic2: ``\{\textit{topic2}\}''
}
\end{displayquote}

\subsection*{\llmbased-Topics Corruption}\par
\begin{displayquote}
{\fontfamily{qcr}\selectfont Following is a title, that represents a theme. Corrupt the title such that the theme could not be easily understood by a human reader. The title must be short and readable. You may make the title vague, metaphorical, or designed to pique curiosity without directly revealing the topic\\\\
Title: \{title\}\\
New Title: 
}
\end{displayquote}\par

\subsection*{LDA Word-Cluster Conversion to \llmbased-Topics Set}\par
\begin{displayquote}
{\fontfamily{qcr}\selectfont Following is a list of words extracted with an LDA model, representing an LDA cluster. Please give a title to the topic this cluster represents\\\\
Cluster words: [\{``, ''.join(words)\}]\\
Title: 
}
\end{displayquote}\par

\subsection*{LLM-based Topics Set Generation}\par
\underline{\textit{Single \llmbased-topics set Generation}}
\begin{displayquote}
{\fontfamily{qcr}\selectfont You are a Holocaust researcher. You will perform the following instructions as best as you can. You will be displayed multiple texts. Please make a list of \{NUM-TOPICS\} unique topics that are common for all of the following texts. Make sure that the topics are general in their description, relevant to the texts, distinct, comprehensive, specific, interpretable, and short.\\
Desired format: \\\\
1. <topic1> \\
2. <topic2> \\
3. <topic3> \\
... \\\\
Text 1: <text1> \\
Text 2: <text2> \\
Text 3: <text3> \\
Text 4: <text4> \\
... \\
Text <N>: <textN> \\
}
\end{displayquote}\par
\underline{\textit{Sets Aggregation}}
\begin{displayquote}
{\fontfamily{qcr}\selectfont 
You will be presented with a set of topic titles. Please choose \{NUM-TOPICS\} distinct titles that best describe the set. Make sure that the topics are distinct, comprehensive, specific, interpretable, and short. \\\\
Desired format: \\
1. <topic1> \\
2. <topic2> \\
3. <topic3> \\
...\\\\

1. <topic1> \\
2. <topic2> \\
3. <topic3> \\
... \\
<N>. <topicN> \\
}
\end{displayquote}\par

%%%%%%%%%%%%%%%%%%%%%%%%%%%%%%%%%%%%%%%%%%%%%%%%

\section{Models and Computations} 

\subsection{LLM Model Versions} \label{app:llm_versions}

Since off-the-shelf LLM are updated by the day, we report the exact model versions used in this work in Table \ref{tab:llm_versions}. 

\subsection{Computational Cost}

During the experimentation stage of our work, we employed different LLM models. To run the models, we have used both the University's GPU infrastructure (mainly used 3 GPUs with memory of 48GB each) and AWS Cloud services (EC2, AWS Bedrock). We report the model versions in \S\ref{app:llm_versions}. The different properties (e.g., number of parameters) of these models can be found online based on the version, if published by developers. Overall, we estimate the computational cost of about 2 weeks of GPU run time.

\begin{table}[h]
\centering
\begin{tabular}{lcc} 
\toprule
\textbf{Developer} & \textbf{Model Family} & \textbf{Version} \\ 
\midrule
OpenAI & GPT & \makecell{\texttt{gpt-4-0125-preview},\\\texttt{gpt-3.5-turbo-0125}, \\\texttt{gpt-4o-mini-2024-07-18}} \\ \hline
Meta & LLaMA & \makecell{\texttt{Meta-Llama-3-8B-Instruct},\\\texttt{Meta-Llama-3-70B-Instruct}} \\ \hline
Mistral & Mistral & \texttt{Mixtral 8x7B} \\ 
\bottomrule
\end{tabular}
\caption{LLM model versions used in this work, grouped by model family}
\label{tab:llm_versions}
\end{table}

%%%%%%%%%%%%%%%%%%%%%%%%%%%%%%%%%%%%%%%%%%%%%%%%

\section{Additional Results} \label{app:additional_results}

\subsection{Judge Model Evaluation}
Table \ref{tab:judge_full} extends Table \ref{tab:judge}. Table \ref{tab:human_to_mean} reports the correlation between human annotations and the mean human score used as the test set, showing that the scores given by the different annotators are highly correlated with the mean score.

\begin{table*}[ht!]
\centering
\begin{tabular}{lc?ccc|ccc|ccc} 
\toprule
\multicolumn{1}{l}{\textbf{Model}} & \multicolumn{1}{c}{\textbf{Quant.}} & \multicolumn{3}{c}{\textbf{Relevance}} & \multicolumn{3}{c}{\textbf{Overlap}} & \multicolumn{3}{c}{\textbf{Interpretability}} \\
 & \multicolumn{1}{l}{} & \multicolumn{1}{c}{Pear.} & \multicolumn{1}{c}{Spear.} & \multicolumn{1}{c}{Kend.} & \multicolumn{1}{c}{Pear.} & \multicolumn{1}{c}{Spear.} & \multicolumn{1}{c}{Kend.} & \multicolumn{1}{c}{Pear.} & \multicolumn{1}{c}{Spear.} & \multicolumn{1}{c}{Kend.} \\
\midrule
GPT 4 & - & \textbf{0.70} & \textbf{0.66} & \textbf{0.52} & 0.89 & 0.86 & 0.74 & 0.72 & 0.63 & 0.47 \\
GPT 3.5 & - & 0.53 & 0.50 & 0.40 & 0.82 & 0.79 & 0.68 & \textbf{0.76} & \textbf{0.73} & \textbf{0.59} \\
GPT 4o Mini & - & 0.62 & 0.61 & 0.48 & \textbf{0.90} & \textbf{0.87} & 0.76 & 0.62 & 0.61 & 0.47 \\
\midrule
LLaMA 3 (8B) & None & 0.44 & 0.45 & 0.37 & 0.88 & 0.85 & 0.76 & 0.67 & 0.54 & 0.43 \\
LLaMA 3 (70B) & 4-bit & 0.39 & 0.48 & 0.40 & 0.31 & 0.24 & 0.22 & 0.16 & 0.29 & 0.22 \\
LLaMA 3 (70B) & None & \underline{0.64} & \underline{0.62} & \underline{0.49} & \underline{\textbf{0.90}} & \underline{\textbf{0.87}} & \underline{\textbf{0.77}} & 0.67 & \underline{0.66} & \underline{0.51} \\
\midrule
Mixtral (8x7B) & 4-bit & 0.44 & 0.43 & 0.34 & 0.88 & 0.83 & 0.72 & 0.73 & 0.65 & 0.50 \\
Mixtral (8x7B) & None & 0.53 & 0.50 & 0.39 & 0.79 & 0.73 & 0.65 & \underline{0.74} & 0.65 & \underline{0.51} \\
\bottomrule
\end{tabular}
\caption{An extension of table \ref{tab:judge}. Showing Pearson \cite{freedman2007statistics}, Spearman \cite{spearman1961proof} and Kendall \cite{kendall1948rank} correlation between LLM and mean human annotations. The best overall model for each measurement is boldfaced and the best open-source alternative is underlined.}
\label{tab:judge_full}
\end{table*}

\begin{table*}
\centering
\begin{tabular}{l?ccc|ccc|ccc} 
\toprule
\multicolumn{1}{l}{} & \multicolumn{3}{c}{\textbf{Relevance}} & \multicolumn{3}{c}{\textbf{Overlap}} & \multicolumn{3}{c}{\textbf{Interpretability}} \\
\multicolumn{1}{l}{} & \multicolumn{1}{c}{Pear.} & \multicolumn{1}{c}{Spear.} & \multicolumn{1}{c}{Kend.} & \multicolumn{1}{c}{Pear.} & \multicolumn{1}{c}{Spear.} & \multicolumn{1}{c}{Kend.} & \multicolumn{1}{c}{Pear.} & \multicolumn{1}{c}{Spear.} & \multicolumn{1}{c}{Kend.} \\
\midrule
Annotator 1 & \textbf{0.93} & 0.67 & \underline{0.58} & \underline{\textbf{0.95}} & \underline{0.93} & \underline{0.89} & \textbf{0.92} & \textbf{0.91} & \textbf{0.8} \\
Annotator 2 & \underline{0.85} & \textbf{0.95} & \textbf{0.89} & - & - & - & - & - & - \\
Annotator 3 & 0.90 & \underline{0.66} & \underline{0.58} & \textbf{0.95} & \textbf{0.96} & \textbf{0.91} & \underline{0.86} & \underline{0.77} & \underline{0.66} \\
Annotator 4 & 0.92 & 0.71 & 0.62 & - & - & - & 0.91 & 0.83 & 0.71 \\
\bottomrule
\end{tabular}
\caption{Correlation of each annotator with the mean human annotation used as the test set. The annotators with max./min. correlation for each metric is boldfaced/underlined, respectively.}
\label{tab:human_to_mean}
\end{table*}

\begin{table*}[ht!]
\centering
\begin{tabular}{m{20em}m{20em}} 
\toprule
\textbf{Original Description} & \textbf{Corrupted Description} \\
\midrule
``Fear of being shot by Germans'' & ``Trepidation Under Teutonic Projectiles'' \\
``Inhumane conditions in the concentration camps''  & ``Unkind States at Encampment Zones''\\
``Disbelief''& ``Dissonant Credence''\\
``Encounter with Russian soldiers''& ``Conflux with Rus Algid Militants''\\
``Russian liberation''& ``Slavic Unshackling''\\
``Discovery of bodies and evidence of mass killings''& ``Unearthed Enigmas: Corporeal Clusters \& Mortality Indices''\\
``Food''& ``Nourishment Alchemization Elements''\\
``Hospitals and medical treatment''& ``Healing Havens and Remedial Maneuvers'' \\
``Red Cross''& ``Crimson Intersection''\\
``Bombings and attacks''& ``Explosive Events and Assaults Unclear''\\
\bottomrule
\end{tabular}
\caption{Examples of \llmbased-topics corruptions generated using GPT4.}
\label{tab:title_corr}
\end{table*}

%%%%%%%%%%%%%%%%%%%%%%%%%%%%%%%%%%%%%%%%%%%%%%%%

\section{\llmbased-Topics Set Generation Systems} \label{app:generation}

Examples of generated \llmbased-topics set for each generation system are shown in Table \ref{tab:validation_examples}.

%%%%%%%%%%%%%%%%%%%%%%%%%%%%%%%%%%%%%%%%%%%%%%%%

\section{Applying \methodname\ on \llmbased-Topic Sets: Extended Results} \label{app:validation_results}

\begin{figure}
    \centering
    \includegraphics[width=0.85\linewidth]{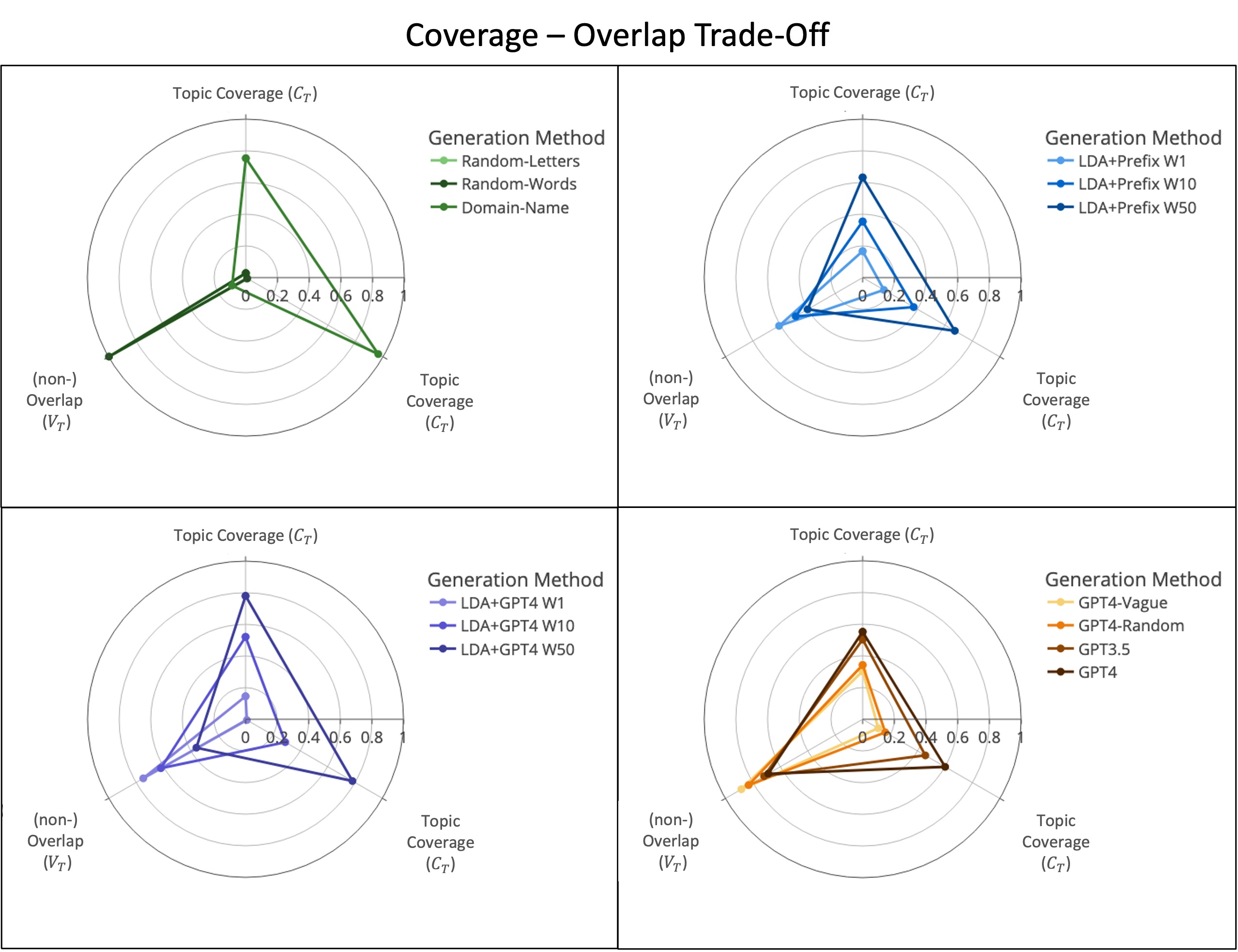}
    \caption{Coverage - Overlap trade-off for all systems, grouped by generation approach.}
    \label{fig:coverage-overlap-all}
\end{figure}

\begin{figure}
    \centering
    \includegraphics[width=0.85\linewidth]{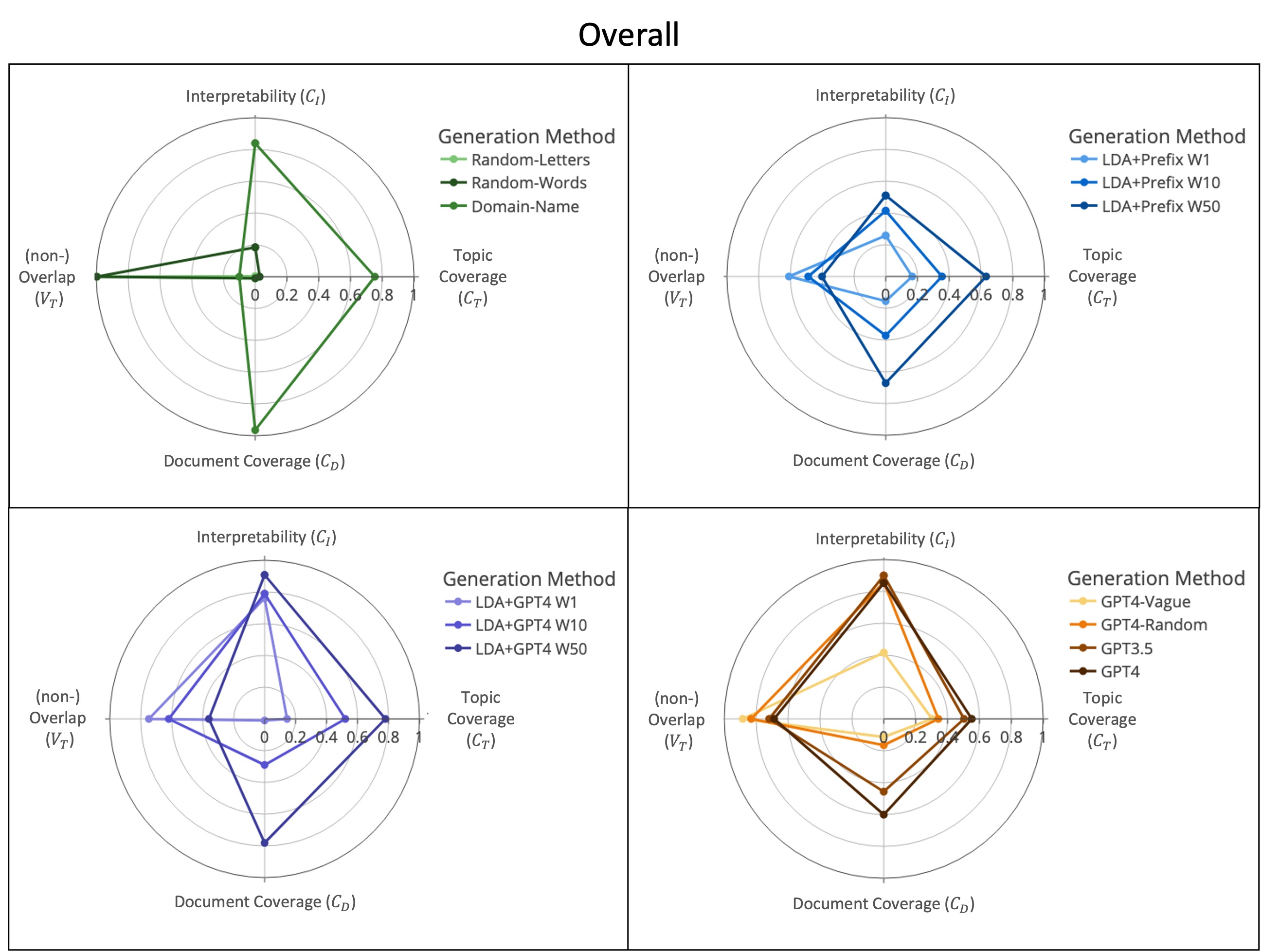}
    \caption{Overall comparison of all aspects other than Inner-Order for all systems. Grouped by generation approach.}
    \label{fig:overall-all}
\end{figure}

\begin{figure}
    \centering
    \includegraphics[width=1\linewidth]{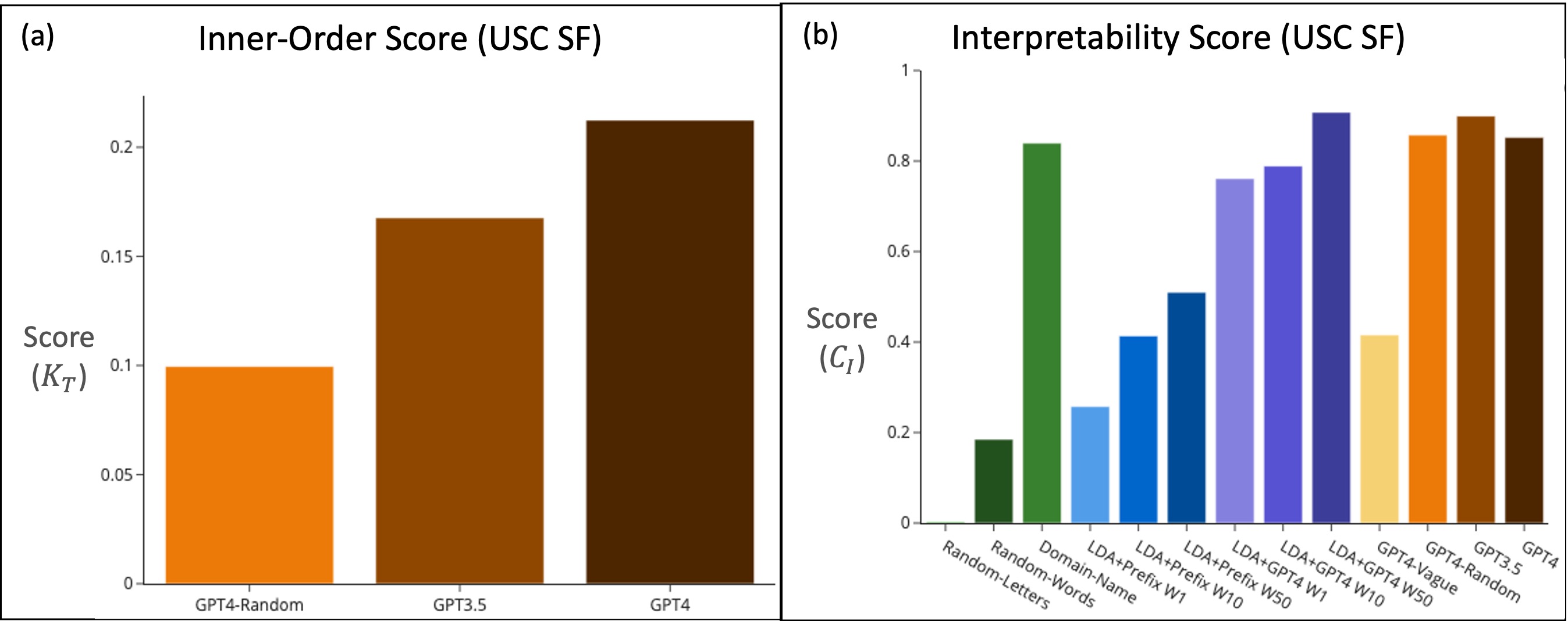}
    \caption{Interpretability and Inner-Order scores for all systems (that participate).}
    \label{fig:interp-rank}
\end{figure}

\begin{figure}
    \centering
    \includegraphics[width=1\linewidth]{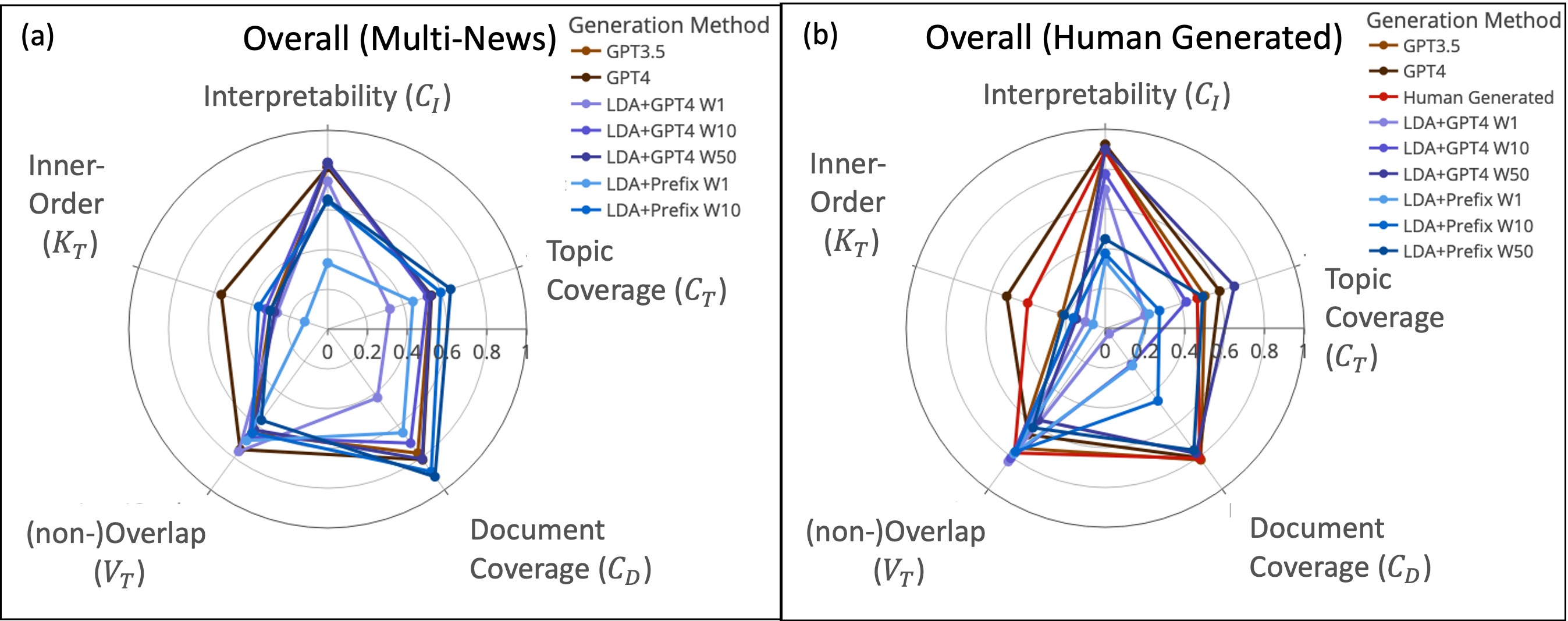}
    \caption{Overall comparison of different generation methods applied to (a) Multi-News dataset and (b) Human Generated Set study and evaluated by \methodname.}
    \label{fig:multi-new_human-baseline_overall_full}
\end{figure}

This section shows a thorough analysis of the results presented in \S\ref{sec:case_study}, further expanding on the trade-offs arising between the different \llmbased-topics sets due to the different generation approaches. Figures \ref{fig:coverage-overlap-all}, \ref{fig:overall-all}, \ref{fig:interp-rank} show the same trade-offs presented in Fig. \ref{fig:validation_aspects}(a-d) based on USC-SF data but extended to include all tested systems. Fig. \ref{fig:multi-new_human-baseline_overall_full} shows the extended overall results for the Multi-News dataset and human-generated set case study. 

\subsection{Coverage-Overlap Trade-Off} 

Fig. \ref{fig:coverage-overlap-all} provides an extended view of the ordering between systems from the perspective of the Coverage and (non-)Overlap aspects.
This view is interesting because it captures a major trade-off in \llmbased-topics set generation. 
To achieve high coverage of the themes in the corpus, one may choose to create very broad sets, that is, containing high-level topics. 
However, it is harder to choose topics such that they are both broad and don't overlap. 
For example, the topic ``Holocaust Experiences'' covers most, if not all, Holocaust-related documents, but any other Holocaust-related topic will also overlap with it.
Therefore, sets of size $N>1$ that include this topic will result in high Coverage scores but low (non-)Overlap scores.
We can see that our methodology successfully captures this trade-off. 

Moreover, we can see from the plots that our expectation about the relative order of the different generation systems holds.
We begin by comparing the Synthetic systems. 
Since Random-Letters and Random-Words generate topics randomly, the resulting topic sets should not contain descriptions that cover the documents nor are overlapping. 
We indeed see that these systems score low on Coverage but high on the (non-)Overlap aspects and surpass all other methods.

Domain-Name utilizes the domain names assigned by SF.
These are intended to describe the entire domain. 
Therefore, most Holocaust-related documents should be covered by it, achieving higher coverage than most systems.
However, since the sets are compiled of the \textit{same} topic, they should maximally overlap.
This is depicted by the resulting Coverage and (non-)Overlap scores.

We move on to examining the LDA-Based methods.
LDA is a widely used and studied topic-modeling approach \cite{stammbach2023re}.
This method produces word distribution clusters representing topics in the corpus.
One of the main drawbacks of LDA is that it produces overlapping clusters. 
This phenomenon is further demonstrated in Table \ref{tab:lda_word_overlap}, showing that the word-overlap between topic clusters extracted from USC-SF domains is high, and more interestingly, it increases as the number of inspected top-$k$ words increases.
In addition, to transform the \ldalike-topic set to \llmbased-topic set, we pick the top-$k$ words from each cluster. 
That is, as we increase $k$, more information is used to describe the topic. 
Therefore, we expect to see that the Coverage of the systems increases as we increase $k$ while the (non-)Overlap decreases.
This is captured by the methodology, as can be seen in the figure.
In addition to that, due to the strong capabilities of LLMs, we expect that LDA+GPT systems will achieve overall higher scores, as can be seen in the figure as well.

Next, we will examine the behavior of the GPT-based methods.
First, due to the strong capabilities of LLMs, we expect that they will surpass all other methods. 
In addition, we expect that newer versions (GPT-4 over GPT-3.5) will achieve higher overall scores since they are claimed to be better models.
We can see this behavior in the plots.

The more interesting case is the comparison of GPT4-Vague and GPT4-Random.
In contrast to the synthetic methods, these correspond to true, interpretable, and Holocaust-related topics.
However, these sets should minimally represent the sets of documents (for being Holocaust-related but not anything specific).
For this reason, we expect that both systems will achieve low Coverage scores, lower than all naturalistic systems but higher than the synthetic ones.
This is captured by our methodology.
Moreover, since these sets are either randomly allocated or corrupted, we expect that they will not overlap, as can also be seen in the figure.

To summarize, from the perspective of the Coverage and (non-)Overlap aspects, we see that our methodology successfully captures the expected order between the generation systems.

\begin{table}[h]
\centering
\begin{tabular}{cc} 
\toprule
\textbf{$k$} & Mean Word Overlap \\ 
\midrule
1 & 0.40 \\ 
10 & 0.55 \\ 
50 & 0.60 \\ 
\bottomrule
\end{tabular}
\caption{Mean number of exact word overlap between pairs of LDA top $k$ words clusters for varying number of words in a cluster. The table shows that the overlap between clusters increases as the number of words in the cluster increases.}
\label{tab:lda_word_overlap}
\end{table}

\subsection{Interpretability Score}

Fig. \ref{fig:interp-rank}(b) shows the Interpretability scores achieved by the different systems. 
The bars in the figure are color-coded, so it will be easier to distinguish between the different system groups. 
Examining the results, we first notice that the methodology successfully captures the low interpretability built into Random-Letters and Random-Words, while human-generated topics (Domain-Name) and systems that employ LLMs (excluding GPT4-Vague) achieve the highest scores.
In the case of GPT4-Vague, the system was specifically designed to output uninterpretable descriptions, which aligns with its low score. 

Furthermore, LLM-based methods achieve comparable scores to humans, aligning with recent claims that LLMs achieve high fluency \cite{yang2023exploring,lai2023multidimensional,jiao2023chatgpt}. 
Additionally, we note that systems based only on LDA (LDA-Prefix) are ranked in the low to mid-score ranges.
This aligns with the main drawback of LDA-based topics, which are difficult to interpret. 

Finally, comparing the LDA-Based method to LLM-based methods, we can see that the methodology successfully captures an improvement in the interpretability score of LDA-Based systems when increasing the number of words, while the score of LLM-based systems remains steady. 
This phenomenon is attributed to the fact that increasing the number of words in an LDA cluster adds substantial useful information, whereas changing the LLM version doesn't necessarily enhance its ability to generate high-quality \llmbased-Topics.

\subsection{Inner-Order Score.}

Fig. \ref{fig:interp-rank}(b) shows the inner-order scores achieved by LLM-based systems.
While LDA-based methods inherently neglect inner ordering, when designing the LLM-based methods, we did not specify any ordering instruction in the generation prompt (see Appendix \ref{app:llm_prompts}). 
In this comparison, we choose to only include systems that were under our control, and for this reason, we choose to only include LLM-based systems.
The results show that the methodology successfully captures the lack of ordering instruction by not significantly surpassing the ``random'' baseline. 
We note, however, that this result may be easily improved by better prompt engineering.

\subsection{Overall Comparison.} 

Fig. \ref{fig:overall-all} depicts an overall comparison of representing systems from each generation group, considering all aspects other than the Inner-Order aspect. 
We notice that both the LDA-based and LLM-based systems, which correspond to applicable systems, achieve high scores on all aspects compared to the baseline methods.
However, it is also hard to tell which model outperforms. Examining the separate metrics, we notice the intricate trade-offs between the systems. While LLM-based methods tend to distribute evenly across aspects, LDA-based methods tend towards higher-level topics, which correspond to high coverage at the expense of non-Overlap and Interpretability. 

%%%%%%%%%%%%%%%%%%%%%%%%%%%%%%%%%%%%%%%%%%%%%%%%

\section{Aggregate Score} \label{app:aggregate_score}

In some use cases, such as quick comparisons between systems or as a reward function for training topic set generation models, an aggregate metric that combines all aspects into a single score is more practical. While the precise formulation may vary across applications, we propose the harmonic mean for the general case due to its simplicity and effectiveness in balancing large and small values, making it well-suited for score averaging. Formally:

\begin{equation}
S(\mathcal{T}_{f},\mathcal{D})\text{=}\frac{|A(\mathcal{T}_{f},\mathcal{D})|}{\sum_{\alpha \in A(\mathcal{T}_{f},\mathcal{D})}\frac{1}{\alpha}}
\end{equation}

where $A(\mathcal{T}_{f},\mathcal{D})$ is the set of aspect scores achieved for the topic set $\mathcal{T}_{f}$ and a set of documents $\mathcal{D}$. Table \ref{tab:agg_score} shows how the different systems fare on the aggregate metric. GPT-4 outperforms all other methods.

\begin{table*}[ht!] 
\centering
\begin{tabular}{l?ccc}
\toprule
\multicolumn{1}{l}{\textbf{Generation Method}} & \multicolumn{3}{c}{\textbf{Dataset}} \\
\multicolumn{1}{l}{} & \multicolumn{1}{c}{USC SF} & \multicolumn{1}{c}{Multi-News} & \multicolumn{1}{c}{Human Baseline} \\
\midrule
Random-Letters & 0.002 & - & - \\
Random-Words & 0.001 & - & - \\
Domain-Name & 0.000 & - & - \\
\midrule
LDA+Prefix W1 & 0.015 & 0.114 & 0.040 \\
LDA+Prefix W10 & 0.179 & 0.372 & 0.186 \\
LDA+Prefix W50 & 0.150 & 0.307 & 0.236 \\
\midrule
LDA+GPT4 W1 & 0.008 & 0.159 & 0.028 \\
LDA+GPT4 W10 & 0.098 & 0.304 & 0.144 \\
LDA+GPT4 W50 & 0.158 & 0.269 & 0.217 \\
\midrule
GPT4-Vague & 0.134 & - & - \\
GPT4-Random & 0.149 & - & - \\
\midrule
GPT3.5 & 0.279 & 0.435 & 0.293 \\
GPT4 & \textbf{0.337} & \textbf{0.524} & \textbf{0.612} \\
\bottomrule
\end{tabular}
\caption{Average aggregate score per generation method and dataset. The best generation method is boldfaced.}
\label{tab:agg_score}
\end{table*}

%%%%%%%%%%%%%%%%%%%%%%%%%%%%%%%%%%%%%%%%%%%%%%%%

\section{Annotation Guidelines} \label{app:annotation_guidelines}

The following includes the annotation guidelines provided for each measurement annotation. Before passing the guidelines to the annotators, a short in-person meeting was conducted where we introduced our research and the specific goals of the annotation session. We introduced the data (Holocaust Testimonies) and discussed its subtilities and sensitivities. Finally, the guidelines and examples were presented and discussed. During the meeting, we have answered any questions raised by the annotators. Each measurement received its own annotation guidelines and was conducted independently: first relevance, then overlap, and finally interpretability. 

\subsection*{Relevance}

{\fontfamily{qcr}\selectfont 
Following is a collection of passages extracted from Holocaust Testimonies. Please read thoroughly each one of the documents. When you finish, you will be shown a passage from the collection along with a set of titles, each title represents a theme. For each passage-title pair, please indicate how relevant is the title to the given passage (0 - not relevant at all, 100 - very relevant).}

\subsection*{Overlap}

{\fontfamily{qcr}\selectfont Attached are the files required to tag the Overlap task. The files include: \\
- A text file containing a collection of passages for annotation (the same passages you have already seen). It is worth opening the file in ``Word'' for ease of reading.\\
- An Excel file containing pairs of titles under the same domain in which you will have to fill in the overlap scores.\\\\ 
The file contains 4 columns: \textit{``domain''}: the label given to the domain by SF; \textit{``topic 1''}, \textit{``topic 2''}: Titles relevant to the domain and that are to be scored; \textit{``score''}: the appropriate score in your opinion from 0 to 100 according to the definition below; \textit{``reasoning''}: your explanation for the score in a short sentence.\\\\
\underline{Task definition}: \\
- Open the text file and read all the passages (you should already be familiar with these passages) \\
- Open the Excel file. For each pair of titles, give a score between 0 and 100 for the degree to which the themes defined by the two titles overlap, in the context of the passages (0 = no overlap at all, 50 = there is a partial overlap, 100 = there is a complete overlap / the titles have the same meaning). }

\subsection*{Interretability}

{\fontfamily{qcr}\selectfont Attached are Excel files containing titles and a text file containing experiences from Holocaust Testimonies. The experiences are the same experiences from previous tasks, but please go through them and read them again. The Excel file contains the titles for labeling. \\\\
\underline{Task definition}: For each title, give a score of 0-100 for the degree to which the title is understandable (75-100 = the theme is understandable, 50-75 = the theme is partially understandable, 25-50 = the theme is poorly understandable, 0-25 = it is not possible to understand what is the intended theme). An understandable title is a title that the theme it induces can be easily understood from the title's text, in the context of the documents. If the theme is clear but not relevant to the documents you have seen, please give a score regardless of the documents and make a note in the ``notes'' column. In addition, you must give a one-sentence explanation of the score. The explanation should be noted in the "explanation" column. \\\\
\underline{Highlights}:\\
- Do you know which parts of the story the title refers to? \\
- Can you find an example in the text that links to the title? \\
- It should be noted that one title may include several topics that are not clearly relevant (in the context of the documents) such that it may not be clear which theme the title describes overall. \\
- Some titles describe features of the theme but do not give a clear and understandable name to the theme. Points should be deducted for this. \\
- Pay attention to the wording, points must be deducted for titles that are not clearly worded. \\
- Points must be deducted in case there is unnecessary information.}

\begin{table*}[!htbp]
\centering
\caption{Examples of segments extracted from the testimonies and the corresponding ontology labels assigned by SF. Speakers are denoted as either ``INT'' for the interviewer or the first letter of the first and last name of the survivor. Note that multiple labels are possible for the same segment.}
\label{tab:seg_examples}
\begin{tabular}{lm{30em}} 
\toprule
\textbf{Labels} & \textbf{Segment} \\
\midrule
\makecell[l]{``Deportation to\\Concentration\\Camps'',\\``Jewish Prayers''} & ``before. INT: When they left-- when-- when they told you to get out of your home, where did they-- SK: We were-- my mother was baking cookies. INT: Yes? SK: We should have for the trip. And they come in, the Gendarmes, but from our same village. We know them. They said, listen, Günczler [NON-ENGLISH], you have to pack your package. You can bring only-- I know the exact details, all. And you have to come up here, in front of the house, five in a row. And I'll come back in 20 minutes, or whatever, and you have to be ready. So my mother put us the clothes on and the food for the kids, whatever we could. And we-- we were waiting there. And they took us for the night to this big [NON-ENGLISH], has a big shul. And there we sit in there. But this is there. I shouldn't repeat it. INT: No, no, it's OK. SK: I will talk about it. Or if you want to start, and then I'll tell you. INT: No, no, no. Just tell me. SK: Now? OK. So when-- so that night, we sit in the shul, everybody and their luggage, and the men saying'' \\ \bottomrule
\makecell[l]{``Deportation to\\Concentration\\Camps'',\\``Forced Marches''} & ``it was all organized by the transport [? Leitung, ?] you know? Everything was seemingly made by our own people. INT: Did you see any Germans? RS: No, no. I didn't. INT: What did you see? How long did the journey take, the walk? RS: Well, it was about four kilometers. INT: Did you arrive at day? What time of day did you arrive? RS: It was night. It was night. INT: Were you marching in the dark? RS: Yes. INT: Were any orders given to you? RS: No, no. INT: Was anybody hit or any punishments given? RS: No. I couldn't see anything. There were Czech gendarmes around, and some SS men. But they didn't touch anybody. INT: What nationality'' \\ \bottomrule
\makecell[l]{``Living conditions'',\\``Protected houses\\(Budapest)''} & ``didn't get along very well. We never did get along very well with her. And all her things were there. And we used all her thing. And we didn't have our own sheets, and our own pillow cases, and our own beddings. But we-- all of us moved, like three-- three or four of us moved into a small room, where she stayed with my-- In the meantime, my sister actually left, too. She was-- she was hiding somewhere. We didn't know where. At one point she disappeared, and my father and I took off the stars, and were looking for her all day long. That was in summer-- must have been July or August. We're looking for her all-- all day long, and then it turned out that she went with-- to yoga teacher. At that time when nobody in Budapest even'' \\ \bottomrule
\end{tabular}
\end{table*}

\begin{table*}[!htbp]
\centering
\caption{Examples of \llmbased-topics sets generated by each system for the domain ``Antisemitism''.}
\label{tab:validation_examples}
\begin{tabular}{m{7em}|c|m{25em}} 
\toprule
\textbf{Generation Method} & \textbf{\makecell{Rank}} & \textbf{Generated Desc.} \\
\midrule
Random-Letters & 1 & DTrHXGOEuctmGDuQd\\  
 & 2 & tHTbUhnToumKgtEedNlkRo\\  
 & 3 & zCPYogMzYgObhMZYiDNexdyZ\\  
 & 4 & lIuAvbK\\  
 & 5 & KkhtVdgzUcAD\\  
 & 6 & qQDlywcXWxvzEhtRjid\\  
 & 7 & JsdcvRfzjTlAYq\\  
 & 8 & ZTPazuWwfFTwnZKoINUU\\  
 & 9 & PloDhuTCp\\  
 & 10 & EZXckfQkRmxGhcS\\  
\midrule
Random-Words & 1 & brachtmema diatomin\\  
 & 2 & garfish obscuring asterisks\\  
 & 3 & select serjeantry vavasories\\  
 & 4 & fathers raylet integrate\\  
 & 5 & restrengthen hoplonemertine\\  
 & 6 & perfectible spondylexarthrosis obtrusiveness\\ 
 & 7 & conventionalism\\  
 & 8 & hotter incoalescence\\  
 & 9 & demulce\\  
 & 10 & underpainting extending circumrotate\\ \midrule
Domain-Name & 1 & antisemitism\\  
 & 2 & antisemitism\\  
 & 3 & antisemitism\\  
 & 4 & antisemitism\\  
 & 5 & antisemitism\\  
 & 6 & antisemitism\\  
 & 7 & antisemitism\\  
 & 8 & antisemitism\\  
 & 9 & antisemitism\\  
 & 10 & antisemitism\\  
\midrule
LDA+Prefix W1 & 1 & The theme defined by the following set of words: ``int''.\\  
 & 2 & The theme defined by the following set of words: ``int''.\\  
 & 3 & The theme defined by the following set of words: ``know''.\\  
 & 4 & The theme defined by the following set of words: ``jewish''.\\  
 & 5 & The theme defined by the following set of words: ``int''.\\  
 & 6 & The theme defined by the following set of words: ``int''.\\  
 & 7 & The theme defined by the following set of words: ``int''.\\  
 & 8 & The theme defined by the following set of words: ``jewish''.\\  
 & 9 & The theme defined by the following set of words: ``int''.\\  
 & 10 & The theme defined by the following set of words: ``int''.\\  
\bottomrule
\end{tabular}
\end{table*}

\begin{table*}[!htbp]
\centering
\begin{tabular}{m{7em}|c|m{25em}} 
\toprule
\textbf{Generation Method} & \textbf{\makecell{Rank}} & \textbf{Generated Desc.} \\
\midrule
LDA+Prefix W10 & 1 & The theme defined by the following set of words: ``int'', ``school'', ``jewish'', ``would'', ``us'', ``know'', ``one'', ``remember'', ``went'', ``time''.\\  
 & 2 & The theme defined by the following set of words: ``int'', ``know'', ``school'', ``jewish'', ``time'', ``jews'', ``jew'', ``one'', ``went'', ``seconds''.\\  
 & 3 & The theme defined by the following set of words: ``know'', ``int'', ``one'', ``school'', ``jewish'', ``remember'', ``would'', ``time'', ``pauses'', ``seconds''.\\  
 & 4 & The theme defined by the following set of words: ``jewish'', ``know'', ``int'', ``used'', ``jews'', ``like'', ``people'', ``school'', ``would'', ``go''.\\  
 & 5 & The theme defined by the following set of words: ``int'', ``jewish'', ``know'', ``like'', ``jews'', ``people'', ``went'', ``said'', ``yes'', ``remember''.\\  
 & 6 & The theme defined by the following set of words: ``int'', ``know'', ``would'', ``school'', ``remember'', ``jewish'', ``one'', ``like'', ``seconds'', ``pauses''.\\  
 & 7 & The theme defined by the following set of words: ``int'', ``going'', ``would'', ``one'', ``bg'', ``english'', ``non'', ``put'', ``went'', ``jew''.\\  
 & 8 & The theme defined by the following set of words: ``jewish'', ``int'', ``know'', ``one'', ``school'', ``seconds'', ``pauses'', ``jews'', ``well'', ``would''.\\  
 & 9 & The theme defined by the following set of words: ``int'', ``know'', ``seconds'', ``pauses'', ``jews'', ``people'', ``jewish'', ``came'', ``would'', ``see''.\\  
 & 10 & The theme defined by the following set of words: ``int'', ``know'', ``school'', ``go'', ``jewish'', ``went'', ``people'', ``us'', ``came'', ``one''.\\  
\midrule
LDA+Prefix W50 & 1 & The theme defined by the following set of words: ``int'', ``school'', ``jewish'', ``would'', ``us'', ``know'', ``one'', ``remember'', ``went'', ``time'', ``yes'', ``go'', ``came'', ``well'', ``jews'', ``children'', ``said'', ``like'', ``even'', ``get'', ``first'', ``home'', ``pauses'', ``think'', ``seconds'', ``people'', ``say'', ``jew'', ``could'', ``got'', ``non'', ``going'', ``much'', ``back'', ``parents'', ``never'', ``day'', ``come'', ``polish'', ``started'', ``called'', ``town'', ``high'', ``always'', ``used'', ``lot'', ``knew'', ``father'', ``boys'', ``german''.\\  
 & 2 & The theme defined by the following set of words: ``int'', ``know'', ``school'', ``jewish'', ``time'', ``jews'', ``jew'', ``one'', ``went'', ``seconds'', ``pauses'', ``yeah'', ``go'', ``children'', ``came'', ``remember'', ``first'', ``said'', ``yes'', ``would'', ``going'', ``us'', ``well'', ``father'', ``say'', ``people'', ``like'', ``antisemitism'', ``ml'', ``non'', ``hitler'', ``war'', ``told'', ``parents'', ``english'', ``years'', ``little'', ``mother'', ``polish'', ``anti'', ``think'', ``german'', ``mean'', ``friends'', ``used'', ``mb'', ``house'', ``thing'', ``old'', ``started''.\\   
\bottomrule
\end{tabular}
\end{table*}

\begin{table*}[!htbp]
\centering
\begin{tabular}{m{7em}|c|m{25em}} 
\toprule
\textbf{Generation Method} & \textbf{\makecell{Rank}} & \textbf{Generated Desc.} \\
\midrule
 & 3 & The theme defined by the following set of words: ``know'', ``int'', ``one'', ``school'', ``jewish'', ``remember'', ``would'', ``time'', ``pauses'', ``seconds'', ``jews'', ``go'', ``went'', ``little'', ``like'', ``jew'', ``really'', ``hl'', ``laughs'', ``father'', ``first'', ``said'', ``came'', ``got'', ``non'', ``child'', ``well'', ``mean'', ``think'', ``say'', ``took'', ``want'', ``could'', ``kind'', ``course'', ``teacher'', ``quite'', ``things'', ``started'', ``us'', ``even'', ``thing'', ``english'', ``yes'', ``knew'', ``come'', ``grade'', ``boy'', ``house'', ``high''.\\  
 & 4 & The theme defined by the following set of words: ``jewish'', ``know'', ``int'', ``used'', ``jews'', ``like'', ``people'', ``school'', ``would'', ``go'', ``non'', ``went'', ``us'', ``jew'', ``one'', ``remember'', ``polish'', ``time'', ``english'', ``war'', ``said'', ``yeah'', ``got'', ``came'', ``lot'', ``seconds'', ``pauses'', ``antisemitism'', ``see'', ``poland'', ``say'', ``even'', ``children'', ``come'', ``always'', ``could'', ``sb'', ``back'', ``mother'', ``well'', ``good'', ``going'', ``little'', ``many'', ``get'', ``called'', ``think'', ``way'', ``took'', ``home''.\\
 & 5 & The theme defined by the following set of words: ``int'', ``jewish'', ``know'', ``like'', ``jews'', ``people'', ``went'', ``said'', ``yes'', ``remember'', ``mother'', ``came'', ``us'', ``would'', ``go'', ``jk'', ``father'', ``well'', ``school'', ``could'', ``fs'', ``polish'', ``time'', ``one'', ``non'', ``little'', ``seconds'', ``pauses'', ``english'', ``think'', ``name'', ``get'', ``yeah'', ``used'', ``see'', ``lot'', ``yiddish'', ``two'', ``war'', ``lived'', ``never'', ``something'', ``really'', ``home'', ``years'', ``oh'', ``tell'', ``say'', ``told'', ``german''.\\  
 & 6 & The theme defined by the following set of words: ``int'', ``know'', ``would'', ``school'', ``remember'', ``jewish'', ``one'', ``like'', ``seconds'', ``pauses'', ``said'', ``go'', ``well'', ``people'', ``came'', ``went'', ``time'', ``yes'', ``jews'', ``used'', ``think'', ``us'', ``going'', ``jew'', ``mother'', ``always'', ``father'', ``things'', ``children'', ``say'', ``got'', ``come'', ``oh'', ``could'', ``little'', ``much'', ``day'', ``first'', ``really'', ``back'', ``knew'', ``home'', ``name'', ``course'', ``see'', ``also'', ``get'', ``two'', ``started'', ``never''.\\  
 & 7 & The theme defined by the following set of words: ``int'', ``going'', ``would'', ``one'', ``bg'', ``english'', ``non'', ``put'', ``went'', ``jew'', ``tape'', ``hiding'', ``well'', ``little'', ``police'', ``day'', ``pauses'', ``take'', ``hit'', ``seconds'', ``course'', ``go'', ``two'', ``thrown'', ``discuss'', ``ways'', ``rocks'', ``among'', ``got'', ``ok'', ``number'', ``next'', ``time'', ``way'', ``think'', ``poland'', ``know'', ``polish'', ``boy'', ``bad'', ``couple'', ``guns'', ``kids'', ``father'', ``killed'', ``laughs'', ``three'', ``say'', ``us'', ``jk''.\\  
 & 8 & The theme defined by the following set of words: ``jewish'', ``int'', ``know'', ``one'', ``school'', ``seconds'', ``pauses'', ``jews'', ``well'', ``would'', ``like'', ``said'', ``people'', ``antisemitism'', ``us'', ``non'', ``time'', ``mother'', ``think'', ``went'', ``go'', ``used'', ``kids'', ``lived'', ``yes'', ``things'', ``little'', ``friends'', ``say'', ``er'', ``name'', ``even'', ``years'', ``german'', ``children'', ``family'', ``father'', ``polish'', ``always'', ``english'', ``came'', ``hl'', ``way'', ``home'', ``called'', ``poland'', ``lot'', ``felt'', ``quite'', ``got''.\\  
\bottomrule
\end{tabular}
\end{table*}

\begin{table*}[!htbp]
\centering
\begin{tabular}{m{7em}|c|m{25em}} 
\toprule
\textbf{Generation Method} & \textbf{\makecell{Rank}} & \textbf{Generated Desc.} \\
\midrule
 & 9 & The theme defined by the following set of words: ``int'', ``know'', ``seconds'', ``pauses'', ``jews'', ``people'', ``jewish'', ``came'', ``would'', ``see'', ``one'', ``well'', ``time'', ``went'', ``said'', ``polish'', ``like'', ``go'', ``us'', ``say'', ``war'', ``remember'', ``could'', ``school'', ``non'', ``yes'', ``many'', ``back'', ``years'', ``english'', ``right'', ``always'', ``going'', ``something'', ``good'', ``poland'', ``first'', ``think'', ``get'', ``started'', ``name'', ``father'', ``yeah'', ``antisemitism'', ``told'', ``called'', ``things'', ``wanted'', ``took'', ``little''.\\  
 & 10 & The theme defined by the following set of words: ``int'', ``know'', ``school'', ``go'', ``jewish'', ``went'', ``people'', ``us'', ``came'', ``one'', ``jews'', ``remember'', ``would'', ``like'', ``said'', ``time'', ``father'', ``going'', ``well'', ``used'', ``back'', ``yes'', ``could'', ``really'', ``pauses'', ``seconds'', ``little'', ``home'', ``mother'', ``non'', ``never'', ``children'', ``say'', ``see'', ``friends'', ``always'', ``antisemitism'', ``get'', ``war'', ``got'', ``house'', ``poland'', ``oh'', ``mean'', ``even'', ``polish'', ``family'', ``many'', ``take'', ``years''.\\  
\midrule
LDA+GPT4 W1 & 1 & Title: Programming and Development\\  
 & 2 & Programming Languages\\  
 & 3 & Understanding or Knowledge Acquisition\\  
 & 4 & Title: Judaism\\  
 & 5 & Numerical Data \& Analysis\\  
 & 6 & Programming and Software Development\\  
 & 7 & Title: Mathematics or Numerical Analysis\\  
 & 8 & Title: Jewish Culture or Religion\\  
 & 9 & Title: Programming and Software Development\\  
 & 10 & ``Given the extremely limited information (only the word ``int''), it's challenging to assign a precise and meaningful title. The word ``int''is commonly used in programming languages to represent ``integer,'' a data type used to store whole numbers. Therefore, without more context or additional words from the cluster, a title could broadly refer to programming or computer science concepts related to numerical data types. Thus, a potential title could be: Title: ``Numeric Data Types in Programming''\\  
\midrule
LDA+GPT4 W10 & 1 & Jewish Education and Experiences\\  
 & 2 & Jewish Education and Community\\  
 & 3 & Title: Educational Reflections in Jewish Contexts\\  
 & 4 & Jewish Cultural and Educational Experiences\\  
 & 5 & Jewish Community and Historical Events\\  
 & 6 & ``School Experiences and Memories in Jewish Communities''\\  
 & 7 & Title: Language and Identity Discussions\\  
 & 8 & Jewish Education and Cultural Discussions\\  
 & 9 & Discussion on Jewish Historical Events and Perspectives\\  
 & 10 & Jewish Community and Education Experiences\\ 
\bottomrule
\end{tabular}
\end{table*}

\begin{table*}[!htbp]
\centering
\begin{tabular}{m{7em}|c|m{25em}} 
\toprule
\textbf{Generation Method} & \textbf{\makecell{Rank}} & \textbf{Generated Desc.} \\
\midrule
LDA+GPT4 W50 & 1 & Title: Memories of Jewish Life and Education\\  
 & 2 & Jewish Life and Antisemitism in Historical Context\\  
 & 3 & Jewish Education and Childhood Memories\\  
 & 4 & Jewish Experience and Antisemitism in Poland\\  
 & 5 & Jewish Life and Heritage Through Personal Narratives\\  
 & 6 & Title: ``Personal Memories and Cultural Reflections''\\  
 & 7 & Title: Conversational Narratives and Conflicts\\  
 & 8 & Jewish Identity and Experiences Amid Antisemitism\\  
 & 9 & Jewish Experience and Antisemitism in Poland\\  
 & 10 & Jewish Family and Antisemitism in Poland\\ 
\midrule
GPT3.5 & 1 & Anti-Semitism in Schools and Education\\  
 & 2 & Segregation and Discrimination in Jewish Communities\\  
 & 3 & Jewish Identity and Pride\\  
 & 4 & Fear and Violence against Jews\\  
 & 5 & Holocaust Survivors' Experiences\\  
 & 6 & Prejudice and Stereotypes\\  
 & 7 & Impact of Nazi Ideology on Education\\  
 & 8 & Discrimination in Schools\\  
 & 9 & Jewish Family and Community Support\\  
 & 10 & Anti-Semitic Propaganda and Indoctrination\\  
\midrule
GPT4 & 1 & Antisemitism\\  
 & 2 & Jewish Education\\  
 & 3 & Jewish Community Life\\  
 & 4 & Personal Experiences of Discrimination\\  
 & 5 & Impact of Nazi Policies\\  
 & 6 & Jewish-Gentile Relations\\  
 & 7 & School Experiences\\  
 & 8 & Family Dynamics\\  
 & 9 & Resistance and Survival Strategies\\  
 & 10 & Post-War Experiences\\  
\midrule
GPT4-Vague & 1 & Anisdeitsm\\  
 & 2 & Hebraic Pedagogy Enigmas\\  
 & 3 & Judaic Communal Existence\\  
 & 4 & Experiential Encodings of Differential Treatment\\  
 & 5 & Policy Influence of N-Axis Entities\\  
 & 6 & JewGent Nexus Dynamics\\  
 & 7 & Educational Episodes\\  
 & 8 & Kinetic Household Constructs\\  
 & 9 & Defiance and Endurance Tactics\\  
 & 10 & Ex-Combat Aftermaths\\  
\bottomrule
\end{tabular}
\end{table*}

\begin{table*}[!htbp]
\centering
\begin{tabular}{m{7em}|c|m{25em}} 
\toprule
\textbf{Generation Method} & \textbf{\makecell{Rank}} & \textbf{Generated Desc.} \\
\midrule
GPT4-Random & 1 & Survival Strategies\\  
 & 2 & Encounters with Local Populations\\  
 & 3 & Smuggling and Black Market\\  
 & 4 & Violence and Persecution\\  
 & 5 & Daily Routine\\  
 & 6 & Immigration and Resettlement\\  
 & 7 & Ghettoization\\  
 & 8 & Post-War Migration\\  
 & 9 & Curfews\\  
 & 10 & Forced Labor\\ 
\bottomrule
\end{tabular}
\end{table*}

\end{document}